%% file: paper.tex
\documentclass{qvereport}
\usepackage{float}
\usepackage{booktabs}
\usepackage{tabularx}
\usepackage{array}
\usepackage{makecell}
\usepackage{amsmath}
\usepackage{amssymb}

\newcolumntype{Y}{>{\centering\arraybackslash}X}

\newcommand{\rowlab}[1]{\rotatebox{90}{\scriptsize\sffamily #1}}
\newcolumntype{R}{>{\centering\arraybackslash}m{3.5mm}}
\newcolumntype{S}{>{\centering\arraybackslash}m{\dimexpr\textwidth-5.5mm\relax}}

\title{\textbf{Instruction-Based Video Editing by Repurposing an Image Editing Model}}

\author[1]{Yunpeng Bai}
\author[2]{Yossi Gandelsman}
\author[2]{Micha\"el Gharbi}
\author[1]{Qixing Huang}   

\affiliation[1]{UT Austin}
\affiliation[2]{Reve}

\input{sections/abstract}
\checkdata[Project Page]{\url{https://yunpeng1998.github.io/Qwen-Video-Edit-Page/}}
\checkdata[Code]{\url{https://github.com/yunpeng1998/Qwen-Video-Edit}}
\checkdata[Model]{\url{https://huggingface.co/yunpeng1998/Qwen-Video-Edit}}
\correspondence{\email{byp215@gmail.com}}

\begin{document}

\maketitle

\begin{figure}[H]
  \centering
  \includegraphics[width=0.2450\textwidth]{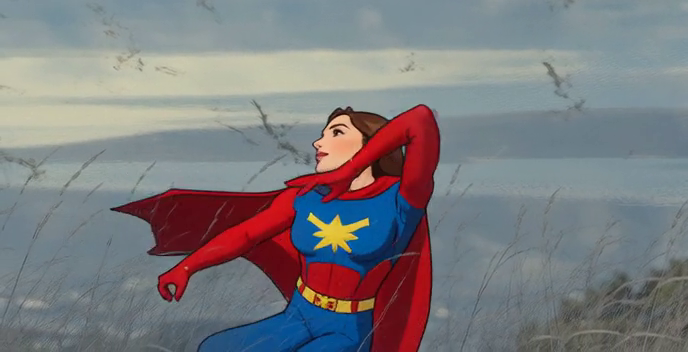}\hfill
  \includegraphics[width=0.2450\textwidth]{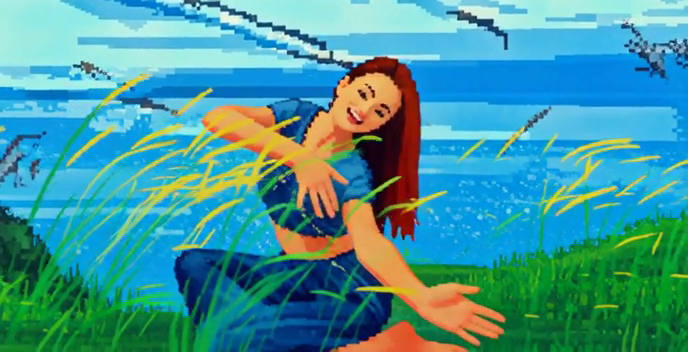}\hfill
  \includegraphics[width=0.2450\textwidth]{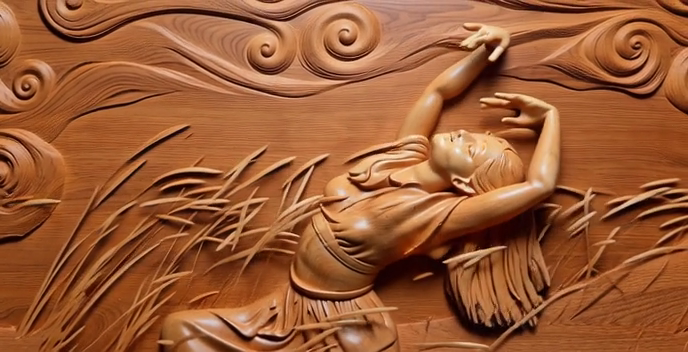}\hfill
  \includegraphics[width=0.2450\textwidth]{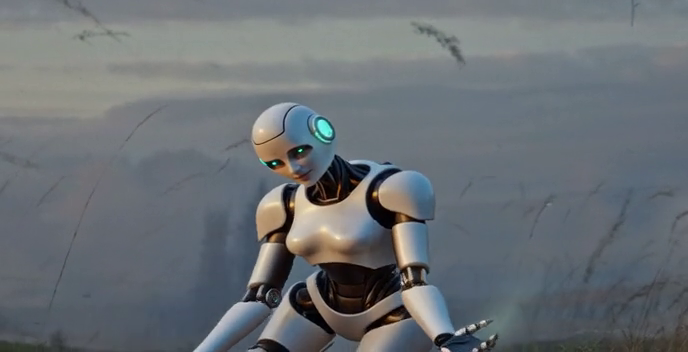}\\[2pt]
  \includegraphics[width=0.2450\textwidth]{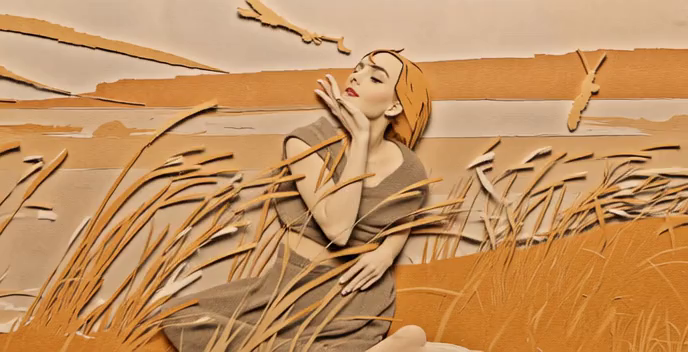}\hfill
  \includegraphics[width=0.2450\textwidth]{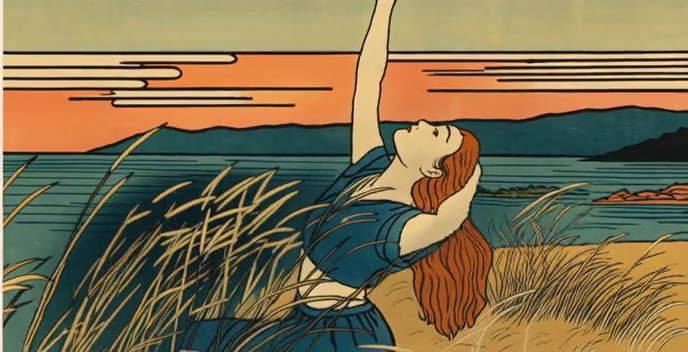}\hfill
  \includegraphics[width=0.2450\textwidth]{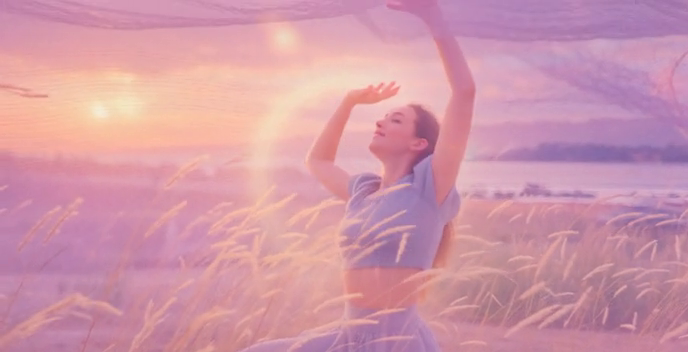}\hfill
  \includegraphics[width=0.2450\textwidth]{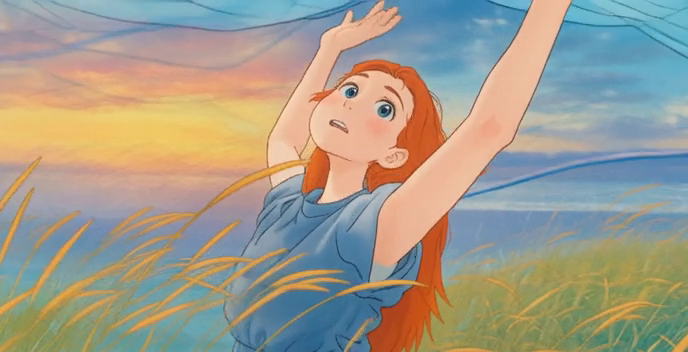}
  \caption{Frames from a \emph{single long video} edited by Qwen-Video-Edit chunk by chunk, each chunk with a different instruction. A pretrained \emph{image} editing transformer operates directly on video-VAE latents; no video-pretrained editing backbone is used.}
  \label{fig:teaser}
\end{figure}

\input{sections/introduction}
\input{sections/related}
\input{sections/empirical}
\input{sections/method}
\input{sections/experiments}
\input{sections/conclusion}

\bibliographystyle{cite}
\bibliography{main}

\end{document}

%% file: sections/abstract.tex
\abstract{
Instruction-based video editing is commonly built on video-pretrained
generative backbones: a video diffusion transformer is adapted, at
considerable cost, to condition on a source video and an editing
instruction. In this report we explore a different route and show that a
strong instruction-based \emph{image} editing model can edit videos by
operating directly on \emph{video-VAE latents}. Starting from
Qwen-Image-Edit, we arrange the latent frames of a Wan~2.1 video VAE as
tiles of one large virtual image, reuse the editor's image positional
encoding for every tile, and bridge the two latent spaces with a pair of
lightweight input/output projections warm-started from the editor's own
patchify and unpatchify layers, so that at initialization a (static) video
is embedded exactly as an image the model already understands. The whole
system is then fine-tuned on the public Ditto-1M editing triplets, and a
few denoising steps of Wan~2.2 serve as an optional temporal enhancer. We
motivate the design with a chain of zero-training observations: the stock
image editor already edits a video presented as a contact sheet; it is
indifferent to whether the sheet's tokens come from one joint encode or
from per-frame encodes stitched in latent space; and it even edits genuine
video latents zero-shot to a clearly recognizable degree, leaving
fine-tuning only a fidelity gap to close. Our
results suggest that, despite the large investment in training video latent
spaces, per-frame video latents remain close enough to the image domain
that mature image editing priors transfer with minimal adaptation.
}

%% file: sections/introduction.tex
\section{Introduction}
\label{sec:intro}

Instruction-based video editing asks a model to modify a source video
according to a free-form textual instruction---restyling a scene, replacing
a subject, changing weather or materials---while preserving the content and
motion that the instruction does not touch. The dominant recipe adapts a
video-pretrained generative backbone to this task: a video diffusion
transformer is conditioned on the source video and the instruction, and
fine-tuned on paired editing data~\citep{zhang2025instructvedit, yu2025veggie,
zi2025senorita, ju2025editverse}. Because large-scale paired video editing
data does not occur naturally, much of the recent progress has in fact been
progress on \emph{synthetic data}: Ditto-1M builds one million editing
triplets with an image-editor-guided in-context video generator and trains
Editto on them~\citep{bai2026scaling}, and JoyAI-Video-Edit pushes the
paradigm to real-time streaming editing with an autoregressive diffusion
backbone~\citep{xiao2026joyai}. All of these systems, however, inherit a
video generation model first and teach it to edit second.

Meanwhile, instruction-based \emph{image} editing has matured much faster.
Models such as InstructPix2Pix~\citep{brooks2023instructpix2pix}, FLUX.1
Kontext~\citep{batifol2025fluxkontext}, Qwen-Image~\citep{qwen_image}, and
Gemini 2.5 Flash Image~\citep{nanobanana} follow nuanced instructions,
perform precise local edits, and preserve identity---capabilities that
video editors are still catching up to. This asymmetry raises a natural
question: rather than teaching a video generator to edit, \emph{can we
teach a mature image editor to handle video}?

The main obstacle appears to be representation. Modern video generators do
not operate on pixels but on the latents of a causal 3D VAE (e.g.\ the
Wan~2.1 VAE~\citep{wan2025}) that compresses a video spatially and
temporally; an image editor has never seen such latents, and its own VAE
encodes single frames only. Our key empirical finding is that this obstacle
is much smaller than it looks. We arrive at it in three steps
(\cref{sec:empirical}). First, a stock image editor already edits videos
presented as an \emph{image grid}: tiling frames into a contact sheet and
editing the sheet with Qwen-Image-Edit applies the instruction to every
frame with surprising cross-frame consistency. Second, the editor is
indifferent to \emph{how the grid was encoded}: encoding each frame
separately with the image VAE and stitching the per-frame latents together
in latent space---with each tile keeping the positional coordinates it
would have had inside the full sheet---produces the same editing behavior.
The transformer therefore does not need one joint pixel-space canvas; it
needs tokens with image-like statistics and consistent positions. Third,
per-frame video-VAE latents are close cousins of exactly such tokens:
substituting Wan~2.1 latents for the stitched image latents---through
projections that merely mimic the editor's own patchify layers, with no
training at all---already produces clearly recognizable instruction-driven
edits. What remains between an image editor and a video editor is a
fidelity gap in per-tile token statistics, not a domain gap.

Concretely, we bridge Qwen-Image-Edit and the Wan~2.1 latent space with two
lightweight projections that replace the editor's patchify and unpatchify
layers and are \emph{warm-started from their weights}, so that at
initialization video latent frames are embedded exactly the way the editor
embeds its own image latents (\cref{sec:method}). Positional encoding
treats the latent frames of both the noisy target and the source video as
tiles of one large virtual image---the same treatment the editor's RoPE
applies to a contact sheet---and the multimodal text branch receives the
source video as a literal contact sheet, keeping the prompt interface
unchanged. We fine-tune the projections and the transformer (LoRA or full
fine-tuning) on Ditto-1M~\citep{bai2026scaling} with the standard flow
matching objective, and optionally apply a few SDEdit-style denoising steps
of Wan~2.2~\citep{wan2025} as a temporal enhancer, following
Ditto's recipe. Long videos are edited chunk by chunk with per-chunk
instructions.

Our contributions are:
\begin{itemize}
    \item A chain of zero-training observations showing that an
    instruction-based image editor extends from pixel-space image grids to
    per-frame latent grids with no quality loss, isolating positional
    treatment---not joint encoding---as the operative ingredient.
    \item A minimal recipe that connects an image editing transformer to a
    video VAE: two warm-started projections, grid positional encoding over
    latent frames, and a contact-sheet prompt, requiring no video-pretrained
    backbone.
    \item An open-source instruction-based video editing system trained on
    public data (Ditto-1M).
\end{itemize}

%% file: sections/related.tex
\section{Related Work}
\label{sec:related}

\subsection{Instruction-Based Image Editing}

Instruction-based image editing has evolved from early proof-of-concept
systems into some of the most capable tools in visual generation.
InstructPix2Pix~\citep{brooks2023instructpix2pix} established the paradigm
of fine-tuning a text-to-image diffusion model~\citep{rombach2022ldm} on
synthetically generated (source image, instruction, edited image) triplets.
Subsequent models scaled both the data and the backbone: FLUX.1
Kontext~\citep{batifol2025fluxkontext}, Step1X-Edit~\citep{liu2025step1x},
Qwen-Image~\citep{qwen_image}, and Gemini 2.5 Flash
Image~\citep{nanobanana} accept text together with reference images,
perform targeted local edits, and maintain subject identity across turns
within a single architecture. Recent work further refines controllability
and perceptual quality, e.g.\ continuous camera-parameter control and
preference-aligned post-training~\citep{qin2025camedit, li2026hpedit}.
Architecturally, these editors are diffusion
transformers~\citep{peebles2023dit} trained with flow
matching~\citep{lipman2022flowmatch, liu2025flow}: the source image is
VAE-encoded, patchified into tokens, and concatenated with the noisy target
tokens, with positional encoding distinguishing the two streams. Our work
takes such an editor---Qwen-Image-Edit~\citep{qwen_image}---as a frozen
starting point and asks how little must change for it to edit videos. Prior
work has also used image editors as \emph{components} of video pipelines,
most notably to edit keyframes that guide a video
generator~\citep{bai2026scaling, yu2025veggie, wu2025insvie}; in contrast,
we use the image editor as the video editor itself.

\subsection{Instruction-Based Video Editing}

\textbf{Inversion-based and training-free methods.}
Early video editing avoided paired data via per-video optimization or
inversion. Tune-A-Video~\citep{wu2023tune} fine-tunes a text-to-image model
on the single input video; TokenFlow~\citep{tokenflow2023},
FateZero~\citep{qi2023fatezero}, and Pix2Video~\citep{ceylan2023pix2video}
propagate inverted features across frames; Video-P2P~\citep{liu2024videop2p},
StableVideo~\citep{chai2023stablevideo}, CoDeF~\citep{ouyang2024codef}, and
AnyV2V~\citep{ku2024anyv2v} refine this recipe with attention control,
layered representations, or image-to-video propagation. These methods
require no editing dataset but are slow per video and fragile under complex
motion.

\textbf{Feed-forward methods and synthetic datasets.}
End-to-end editors are bounded by data. InstructVid2Vid~\citep{qin2024instructvid2vid}
and EffiVED~\citep{zhang2024effived} synthesized training pairs with
one-shot-tuned editors; VEGGIE~\citep{yu2025veggie} and
InsViE~\citep{wu2025insvie} edit a keyframe and propagate it with an
image-to-video model; Se\~norita-2M~\citep{zi2025senorita} assembles a
large corpus from a suite of task-specific expert models;
InsV2V~\citep{cheng2023insv2v}, InstructVEdit~\citep{zhang2025instructvedit},
DreamVE~\citep{xia2025dreamve}, and UniVideo~\citep{wei2025univideo}
explore related data and unification strategies, and
EditVerse~\citep{ju2025editverse} unifies image and video editing through
in-context learning. Ditto~\citep{bai2026scaling} scales this line to one
million triplets with a single image-editor-guided in-context video
generator~\citep{jiang2025vace} plus a VLM curation agent, and trains the
Editto model on the result. Beyond offline editing,
JoyAI-Video-Edit~\citep{xiao2026joyai} targets real-time open-ended editing
with an autoregressive diffusion backbone, alongside other streaming
systems~\citep{wang2026liveedit, zhao2026sana, feng2025streamdiffusionv2}.
Common to all of these editors is a video-pretrained generative backbone,
typically inherited from large text-to-video
models~\citep{wan2025, kong2024hunyuanvideo, yang2024cogvideox,
blattmann2023svd}, whose role is to supply temporal priors in the video
VAE's latent space. Our work uses the same public dataset (Ditto-1M) but
removes the video backbone entirely: the temporal dimension is handled by
treating video latent frames as tiles of an image grid, and all generative
capability comes from an image editing model. We use a few denoising steps
of Wan~2.2~\citep{wan2025} only as an optional post-hoc enhancer, following
Ditto's own denoising-enhancement recipe.

%% file: sections/empirical.tex
\section{Empirical Observations}
\label{sec:empirical}

Our method was not designed top-down; it emerged from a chain of
zero-training experiments with the stock Qwen-Image-Edit checkpoint, each
removing one assumption about what the editor needs. We describe them in
the order we ran them. All three use the same interface: $N$ frames sampled
from a video, an editing instruction, and no weight updates whatsoever.

\subsection{Editing a Video as an Image Grid}
\label{sec:empirical_grid}

\begin{figure}[t]
  \centering
  \includegraphics[width=0.492\textwidth]{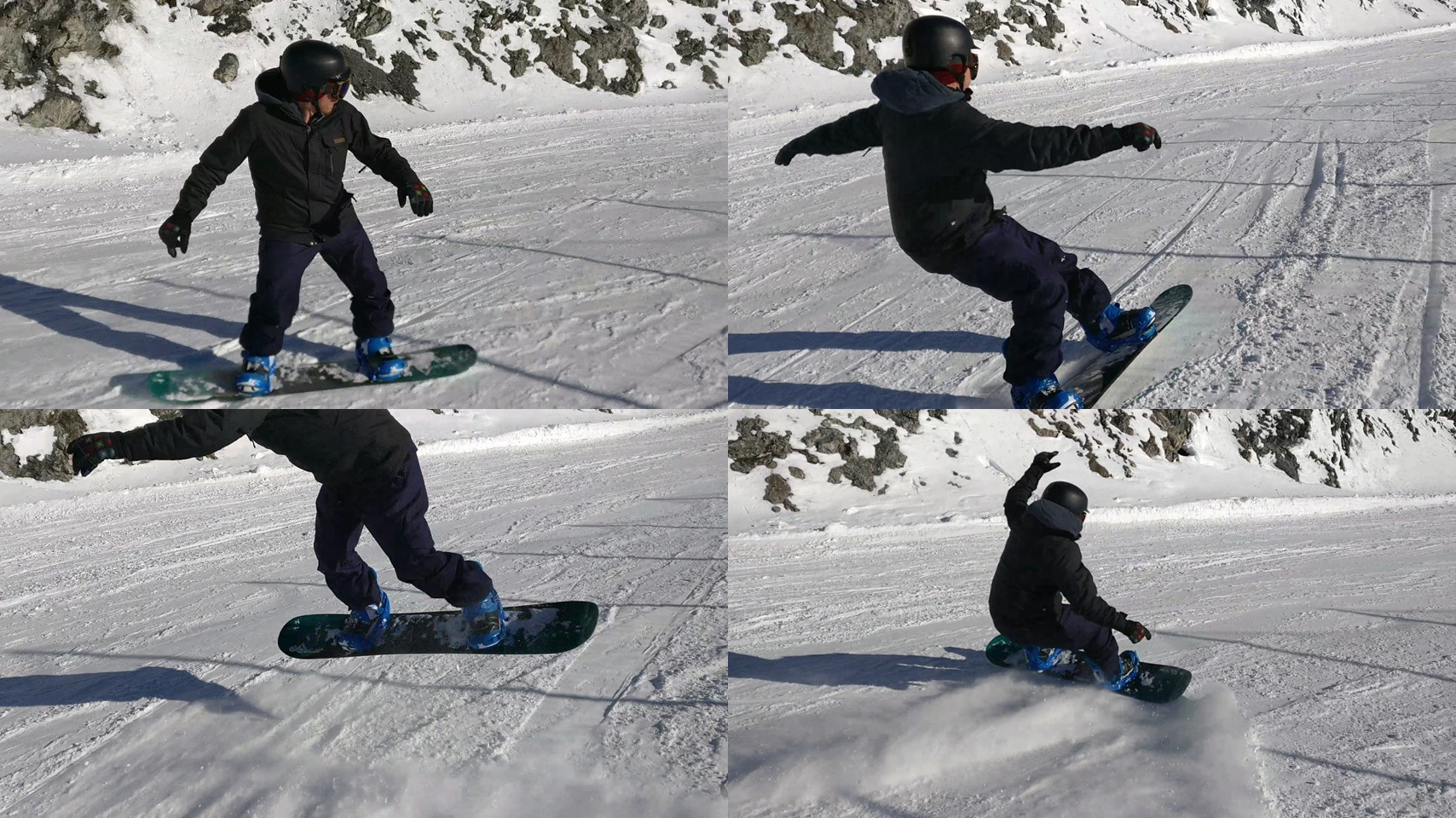}\hfill
  \includegraphics[width=0.492\textwidth]{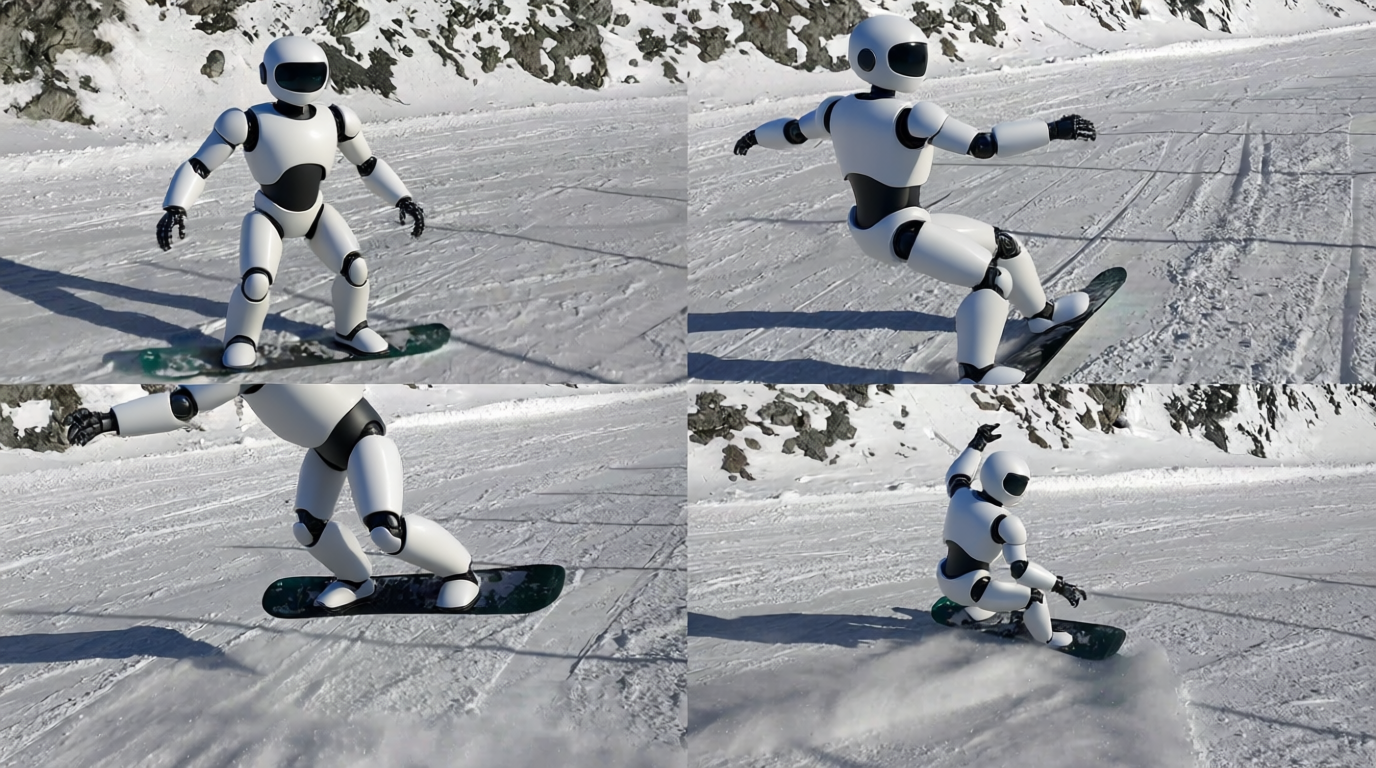}
  \caption{\textbf{Editing a video as an image grid.} Input contact sheet (left) and the output of the unmodified Qwen-Image-Edit (right) for the instruction \emph{``Edit the skier into a robot.''} All frames are edited coherently.}
  \label{fig:empirical_image_grid}
\end{figure}

The first observation is folklore-adjacent but worth stating precisely:
an instruction-based image editor can edit a video presented as a
\emph{contact sheet}. We tile $N{=}r{\times}c$ uniformly sampled frames into
one large image, pass it through the unmodified editing pipeline with the
instruction, and split the result back into frames
(\cref{fig:empirical_image_grid}). The editor applies the instruction to
every tile, and---more surprisingly---keeps the tiles consistent with each
other: subject identity, color palette, and edit strength match across
frames, presumably because tokens from all tiles attend to each other
exactly as tokens within one coherent image would. Two practical
constraints matter. The total canvas must stay near the editor's training
resolution (about $1024^2$ pixels for Qwen-Image-Edit); well beyond it,
global attention degrades and tile structure breaks down.

\subsection{From a Pixel Grid to a Latent Grid}
\label{sec:empirical_latent}

\begin{figure}[t]
  \centering
  \begin{tabular}{@{}R@{\hspace{1.5mm}}S@{}}
    \rowlab{Input} &
    \includegraphics[width=0.1585\linewidth]{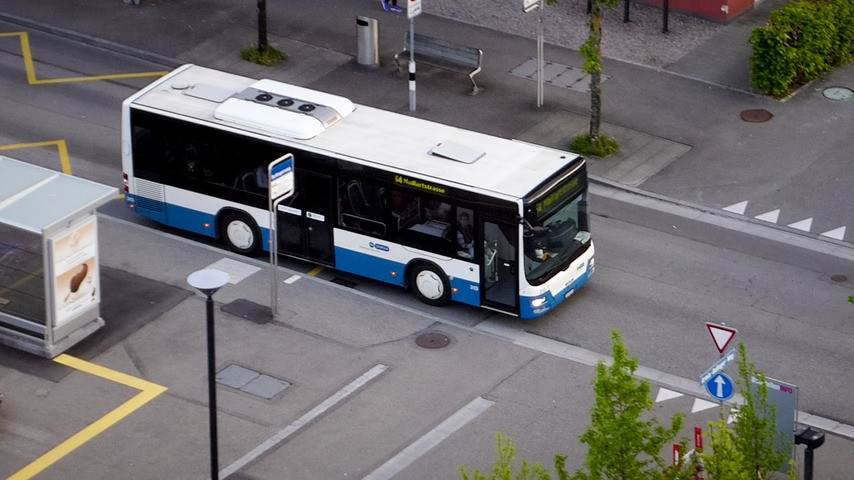}\hfill
    \includegraphics[width=0.1585\linewidth]{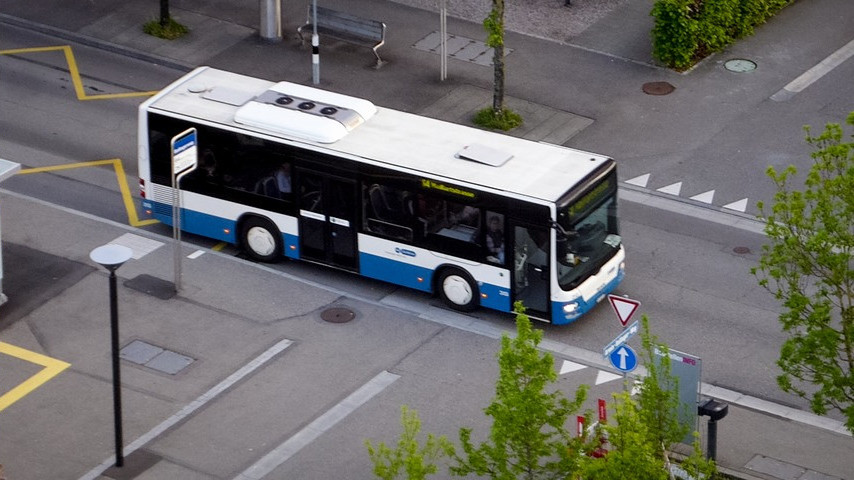}\hfill
    \includegraphics[width=0.1585\linewidth]{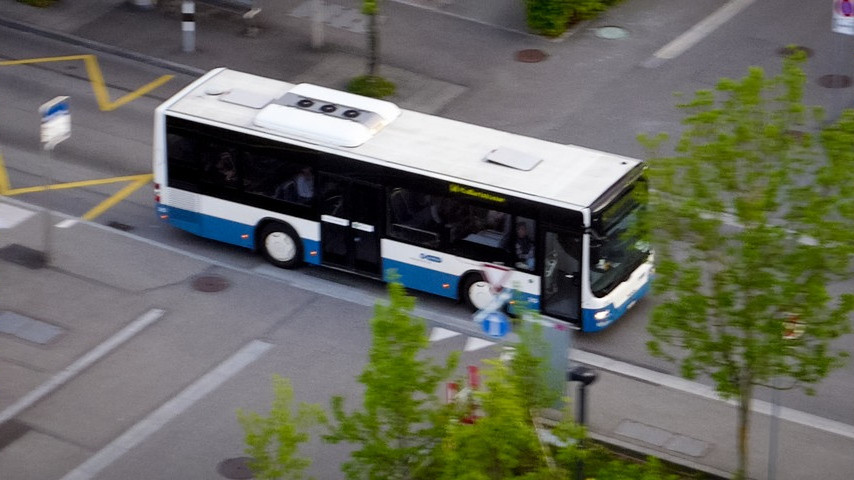}\hfill
    \includegraphics[width=0.1585\linewidth]{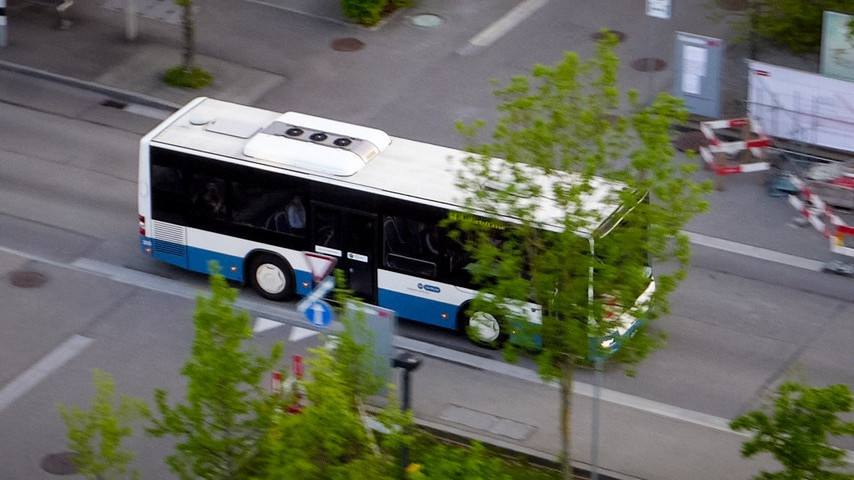}\hfill
    \includegraphics[width=0.1585\linewidth]{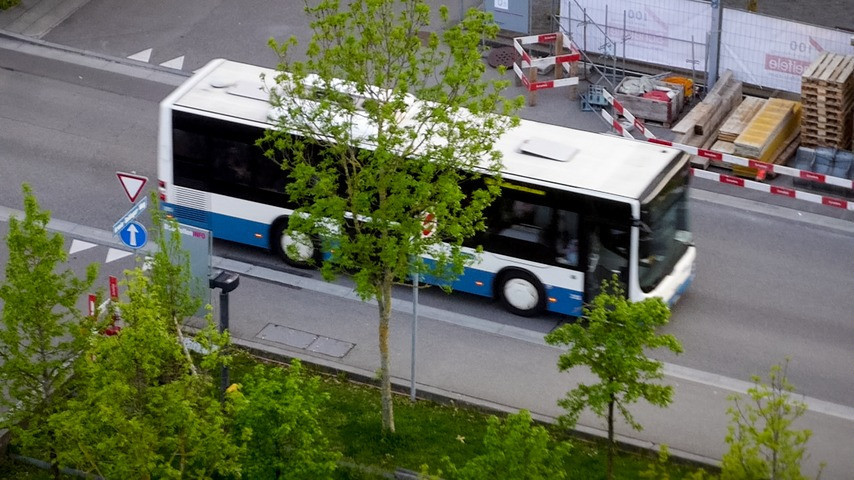}\hfill
    \includegraphics[width=0.1585\linewidth]{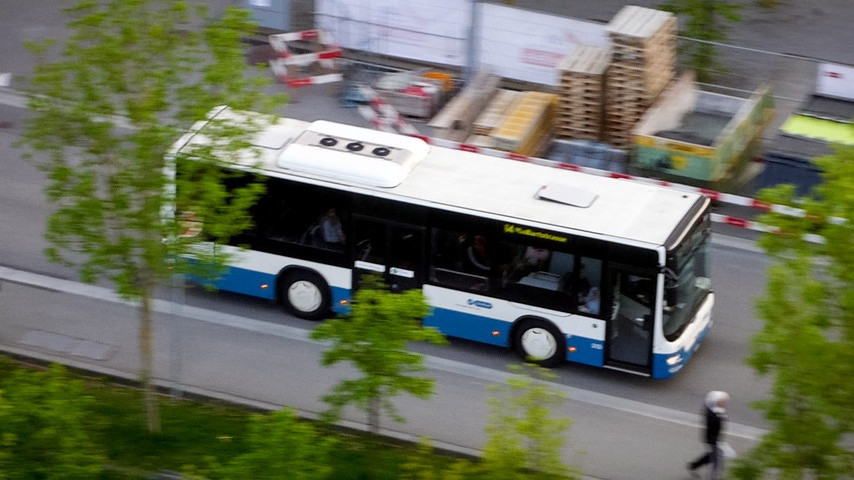} \\[1pt]
    \rowlab{Edited} &
    \includegraphics[width=0.1585\linewidth]{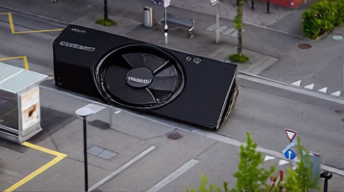}\hfill
    \includegraphics[width=0.1585\linewidth]{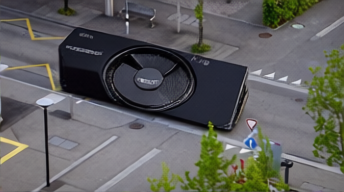}\hfill
    \includegraphics[width=0.1585\linewidth]{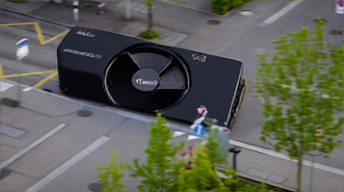}\hfill
    \includegraphics[width=0.1585\linewidth]{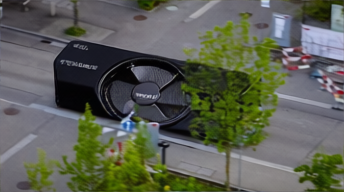}\hfill
    \includegraphics[width=0.1585\linewidth]{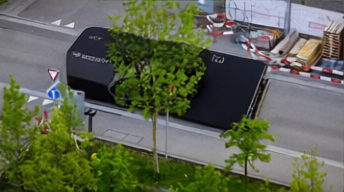}\hfill
    \includegraphics[width=0.1585\linewidth]{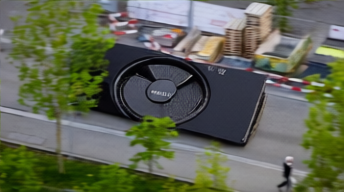}
  \end{tabular}
  \caption{\textbf{Editing a latent grid.} Frames are encoded \emph{separately}, stitched only at the token level (each tile keeping its full-canvas position in a $4{\times}4$ grid), and decoded separately after editing with the instruction \emph{``Turn the bus into a graphics card.''} Shown are six of the sixteen tiles; behavior matches pixel-grid editing---the joint encode is unnecessary.}
  \label{fig:empirical_latent_grid}
\end{figure}

The contact sheet couples the frames \emph{in pixel space}: they are
jointly encoded by the image VAE as one big image. If grid editing
depended on that joint encode, it would be useless for video latents,
which are produced frame by frame. So we removed the coupling: each frame
is encoded \emph{separately} with the image VAE, and the per-frame latents
are concatenated only at the token level. Crucially, each tile's tokens
keep the RoPE coordinates they would have had inside the full canvas---an
explicit frame index plus the tile's spatial offset in the virtual
grid---which is possible because positions in the editor's RoPE are
per-token and attention is permutation-equivariant. Denoising proceeds
with the editor's transformer unchanged, and each edited latent tile is
decoded separately (\cref{fig:empirical_latent_grid}). The result is
indistinguishable from pixel-grid editing. This is the pivotal
observation: the editor does not care whether its tokens came from one
joint encode or from $N$ independent encodes. What it needs is (i)
per-tile token statistics matching its image latents and (ii) consistent
grid positions. Both requirements are properties we can satisfy for
latents that were never produced by its own VAE.

\subsection{From Image Latents to Video Latents}
\label{sec:empirical_video}

\begin{figure}[t]
  \centering
  \begin{tabular}{@{}R@{\hspace{1.5mm}}S@{}}
    \rowlab{Input} &
    \includegraphics[width=0.2440\linewidth]{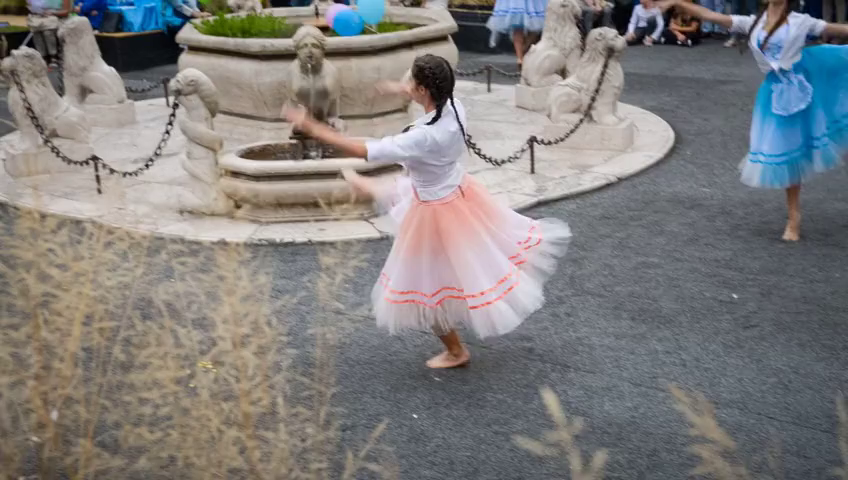}\hfill
    \includegraphics[width=0.2440\linewidth]{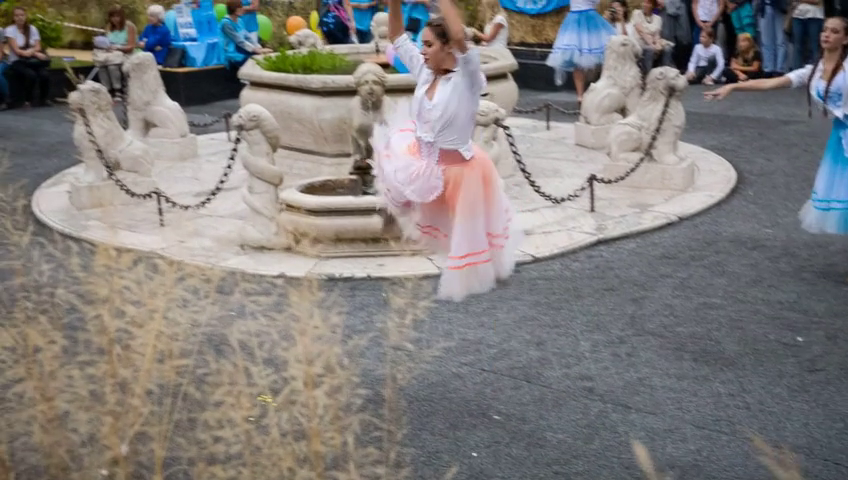}\hfill
    \includegraphics[width=0.2440\linewidth]{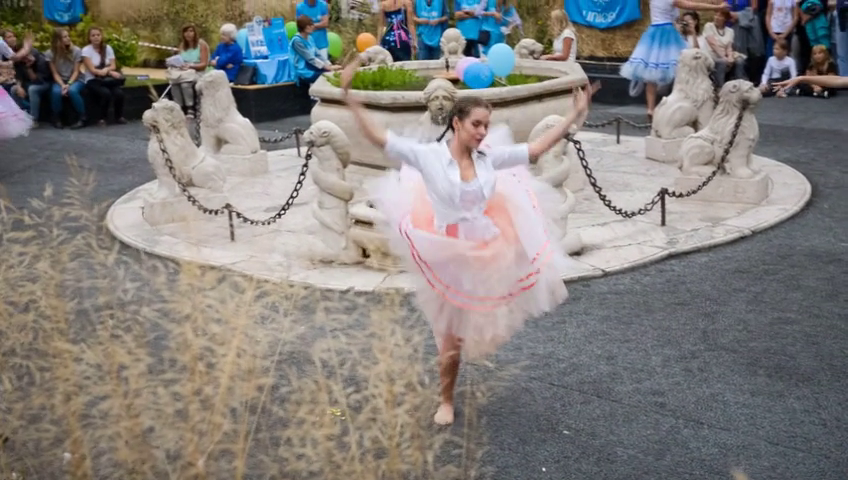}\hfill
    \includegraphics[width=0.2440\linewidth]{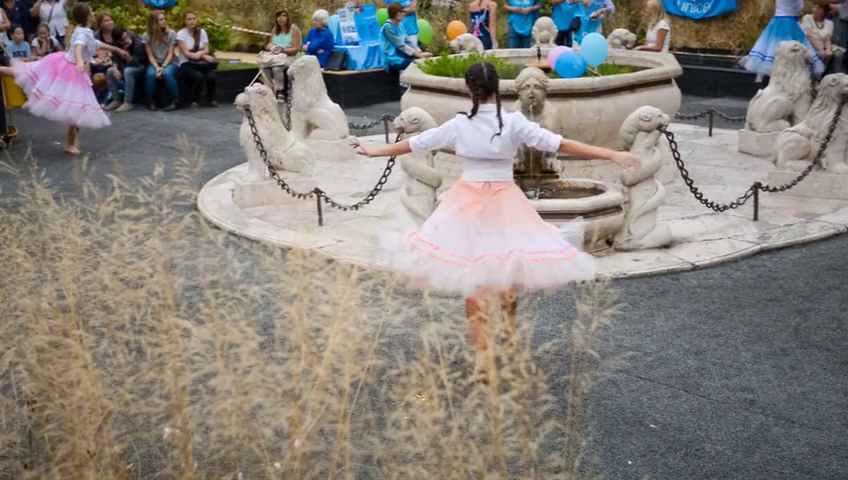} \\[1pt]
    \rowlab{Edited} &
    \includegraphics[width=0.2440\linewidth]{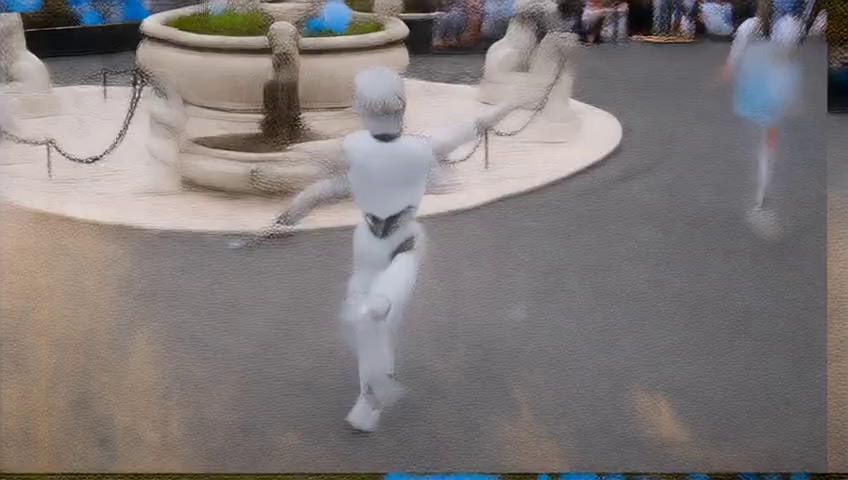}\hfill
    \includegraphics[width=0.2440\linewidth]{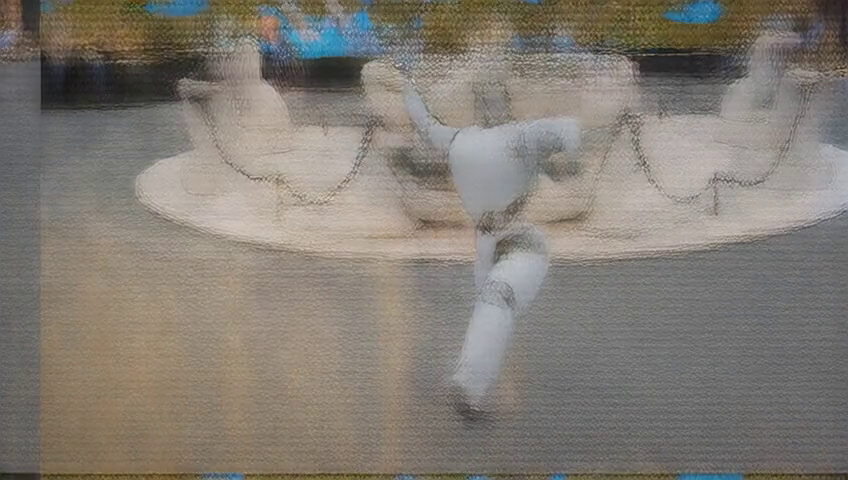}\hfill
    \includegraphics[width=0.2440\linewidth]{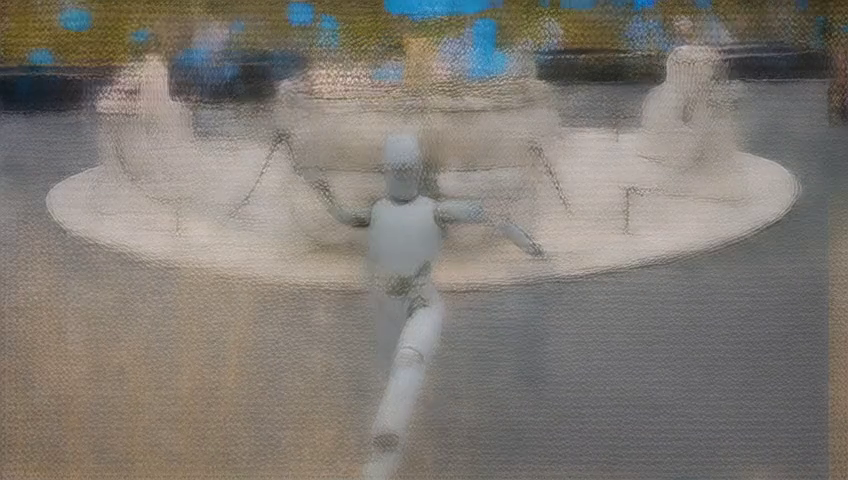}\hfill
    \includegraphics[width=0.2440\linewidth]{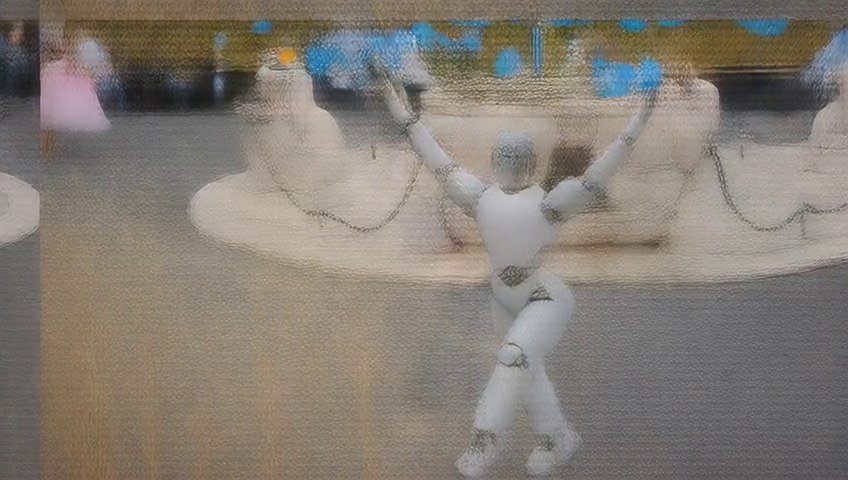}
  \end{tabular}
  \caption{\textbf{Zero-shot editing of video latents.} The \emph{pretrained} Qwen-Image-Edit, with warm-started (untrained) projections and the grid treatment, edits Wan~2.1 video latents directly for the instruction \emph{``Transform the dancing girl in the image into a robot.''} Fidelity degrades, but the edit clearly lands---motivating the lightweight adaptation of \cref{sec:method}.}
  \label{fig:empirical_video_latent}
\end{figure}

The latent grid still uses the editor's own image VAE. The final step
replaces it with a genuine video representation: the per-frame latents of
the Wan~2.1 video VAE, which compress the video $8\times$ spatially and
$4\times$ temporally. These latents have a different channel basis and, for
all but the first latent frame, mix information across time---a priori one
might expect a hard domain gap. Instead we find that the \emph{pretrained}
editor already has traction on them. We feed Wan latent frames through the
warm-started projections of \cref{sec:method_bridge}---which at
initialization simply mimic the editor's own patchify/unpatchify layers,
i.e.\ involve \emph{no training whatsoever}---apply the same grid
treatment, and denoise with the frozen editor.
\Cref{fig:empirical_video_latent} shows the outcome: the instruction is
followed to a clearly recognizable degree, with degraded fidelity but
intact semantics and layout. Zero-shot, an image editing model edits video
latents it has never seen. This is the observation that motivates the rest
of the report: the previous two experiments established that grid
positions, not joint encoding, carry the recipe; this one shows the
remaining mismatch is a matter of per-tile token statistics that the model
can already partially absorb. Closing that gap is therefore a lightweight
fine-tuning problem---not a generative-pretraining problem---and the next
section describes how we set it up.

%% file: sections/method.tex
\section{Method}
\label{sec:method}

\begin{figure}[t]
  \centering
  \includegraphics[width=\textwidth]{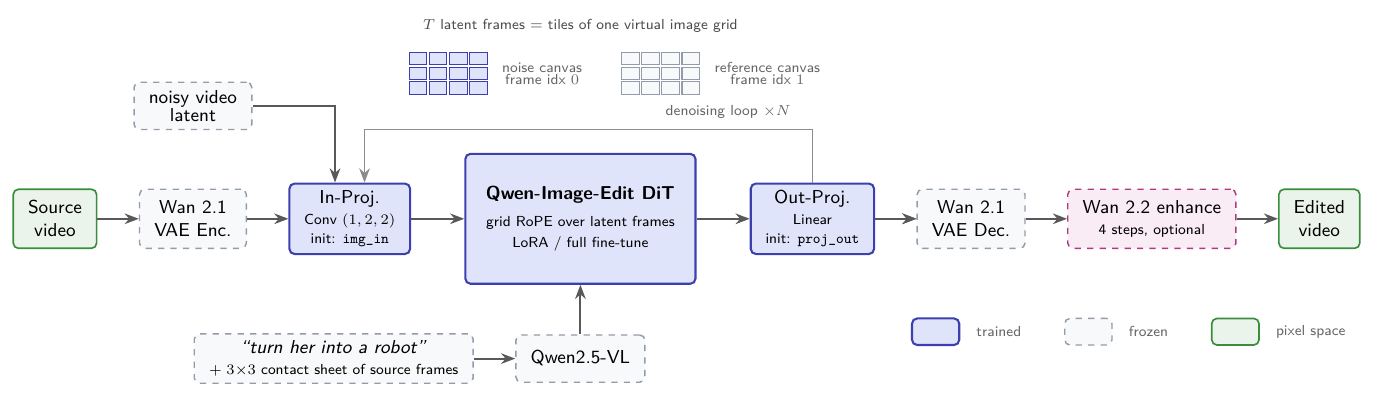}
  \caption{\textbf{Overview.} Video latent frames are embedded as tiles of a virtual image grid and edited by the image editing transformer; only the two projections (warm-started from the editor's own patchify/unpatchify layers) and the transformer weights (LoRA or full) are trained.}
  \label{fig:pipeline}
\end{figure}

Guided by \cref{sec:empirical}, our design goal is to change as little as
possible: keep the image editing transformer, keep its grid positional
treatment, keep its prompt interface, and adapt only the token interface to
the video latent space. \Cref{fig:pipeline} gives an overview.

\subsection{Preliminaries}
\label{sec:method_prelim}

\textbf{Qwen-Image-Edit.}
Qwen-Image-Edit~\citep{qwen_image} is a diffusion
transformer~\citep{peebles2023dit} trained with flow
matching~\citep{lipman2022flowmatch}. Its image VAE produces
16-channel latents at $8\times$ spatial downsampling; the transformer
patchifies them with $2{\times}2$ patches, so one token carries a
$64$-dimensional patch ($16{\times}2{\times}2$) mapped to the model width
by a linear layer $W_{\mathrm{in}}$, and the output layer
$W_{\mathrm{out}}$ maps back to $64$ channels. Positions use a
three-axis RoPE (frame, height, width); for editing, the noisy target
tokens take frame index $0$ and the VAE-encoded source-image tokens take
frame index $1$, with spatial coordinates centered on the canvas. The text
branch is a vision-language model (Qwen2.5-VL~\citep{bai2025qwen25vl})
that sees the instruction together with the source image.

\textbf{Wan 2.1 video VAE.}
The Wan~2.1 VAE~\citep{wan2025} encodes a video of $4k{+}1$ frames into
$k{+}1$ latent frames of $16$ channels at $8\times$ spatial reduction: a
causal 3D encoder whose first latent frame depends only on the first pixel
frame, while each subsequent latent frame summarizes four pixel frames.
Wan~2.2's dense text-to-video models share this VAE, which we exploit for
enhancement (\cref{sec:method_inference}).

\subsection{Video Latents as Multi-Frame Image Latents}
\label{sec:method_bridge}

Let $z \in \mathbb{R}^{16 \times T \times h \times w}$ be the video latent
(source or noisy target) with $T$ latent frames. We treat each latent frame
as if it were an image latent and patchify it exactly as the editor would:

\textbf{Input projection.}
A single 3D convolution with kernel and stride $(1,2,2)$ maps
$z$ to tokens of the model width, acting independently per latent frame.
It is warm-started from the editor's own patchify layer by reshaping
$W_{\mathrm{in}} \in \mathbb{R}^{d \times 64}$ into a
$d{\times}16{\times}2{\times}2$ kernel---numerically exact, so at
initialization a video latent frame is embedded precisely as
Qwen-Image-Edit embeds an image latent with the same values.

\textbf{Output projection.}
A linear layer maps final tokens back to $64$-dimensional
$2{\times}2{\times}16$ patches per latent frame, warm-started from
$W_{\mathrm{out}}$ the same way.

The warm start matters: the projections are the only randomly-placed
components in the system, and initializing them as exact copies of the
patchify/unpatchify pair means training starts from a model whose behavior
on (near-)static content is already the image editor's, rather than from
noise.

\subsection{Positional Treatment and Conditioning}
\label{sec:method_pe}

\textbf{Grid positions.}
Following \cref{sec:empirical_latent}, the $T$ noisy latent frames are
placed as tiles of one virtual canvas in row-major order: every tile keeps
frame index $0$, and its tokens receive the height/width coordinates of
its position inside the $r{\times}c$ grid, exactly as if the canvas had
been encoded as a single image. The $T$ source latent frames form a second,
congruent canvas at frame index $1$. This reproduces, token for token, the
positional layout of the image-grid experiments, and matches the
noise/reference relation ($0$ vs.\ $1$) the editor was pretrained with.

\textbf{Prompt.}
The VL text branch receives the instruction together with a $3{\times}3$
contact sheet of uniformly sampled source frames, so the semantic
conditioning pathway is byte-compatible with image editing: from the
prompt encoder's perspective, the model is still editing one image.


\subsection{Training}
\label{sec:method_training}

We train with the standard flow matching objective on (source video,
instruction, edited video) triplets: the source latent is the condition,
the edited latent is the target, and the model predicts the velocity of
the noisy target tokens. Trainable parameters are the two projections plus
the transformer, either as LoRA on attention and MLP projections or as
full fine-tuning; both VAEs and the VL encoder stay frozen. Because the
token count of a video grid far exceeds a single image, we scale the
flow-matching noise schedule's shift with the actual noise-token count
(the same dynamic-shift rule the editor uses across image resolutions),
during both training-time sampling and inference. Sequence lengths are
kept manageable by the Wan VAE itself: $45$ frames become $12$ latent
frames, i.e.\ a $3{\times}4$ grid of image-latent-sized tiles.

\subsection{Inference and Enhancement}
\label{sec:method_inference}

At inference we denoise the video-latent grid with classifier-free
guidance (an empty negative prompt) and decode with the Wan~2.1 VAE. Finally, following Ditto's denoising-enhancement
recipe~\citep{bai2026scaling}, we optionally re-noise the decoded video to
a small timestep and run a few (four) denoising steps of Wan~2.2's
text-to-video model. Because Wan~2.2 shares the Wan~2.1 VAE, this happens
in the same latent space and costs only a fraction of a full generation;
it removes residual high-frequency flicker while preserving the edit.

%% file: sections/experiments.tex
\section{Experiments}
\label{sec:experiments}

\subsection{Setup}
\label{sec:exp_setup}

\textbf{Data.}
We train on Ditto-1M~\citep{bai2026scaling}, the public dataset of one
million (source video, instruction, edited video) triplets synthesized
with an image-editor-guided in-context video generator and curated by a
VLM agent. Videos are 720p at diverse aspect ratios.

\textbf{Training recipe.}
The base editor is Qwen-Image-Edit; the video VAE is Wan~2.1. We report
two regimes: LoRA (rank $32$ on all attention and MLP projections of the
transformer, learning rate $10^{-4}$) and full fine-tuning of the
transformer (learning rate $10^{-5}$). The two bridging projections are always fully trained.
All runs use bf16, AdamW~\citep{loshchilov2017adamw}, gradient
checkpointing, flow matching with sequence-length-dependent dynamic shift,
and per-GPU batch size $1$ on H100 GPUs.

\textbf{Inference.}
Unless noted, we use $40$ denoising steps, classifier-free guidance scale
$4$ with an empty negative prompt, and four Wan~2.2 enhancement steps.

\subsection{Qualitative Results}
\label{sec:exp_qualitative}

\begin{figure}[!t]
  \centering
  \begin{subfigure}{\textwidth}
    \centering
    \begin{tabular}{@{}R@{\hspace{1.5mm}}S@{}}
      \rowlab{Input} &
      \includegraphics[width=0.2300\linewidth]{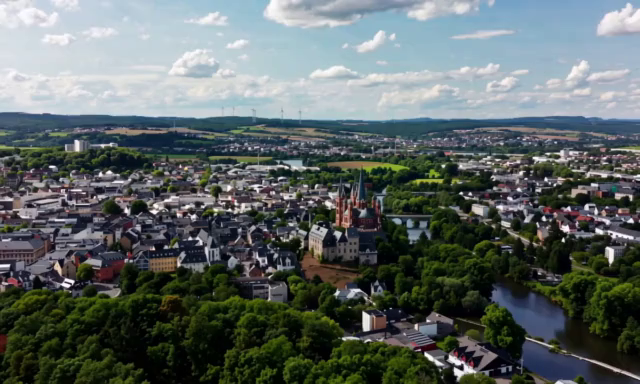}\hfill
      \includegraphics[width=0.2300\linewidth]{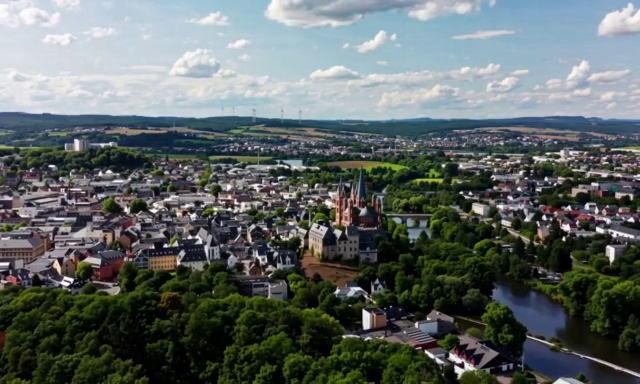}\hfill
      \includegraphics[width=0.2300\linewidth]{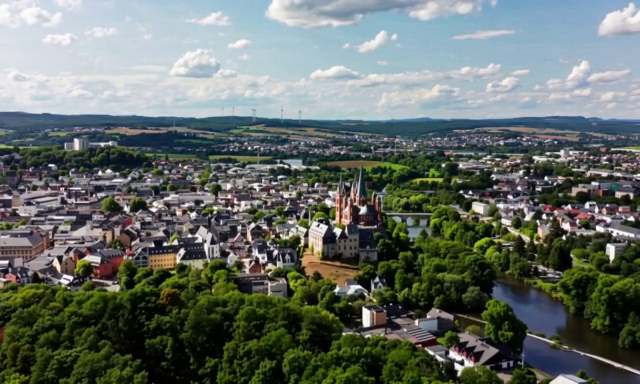}\hfill
      \includegraphics[width=0.2300\linewidth]{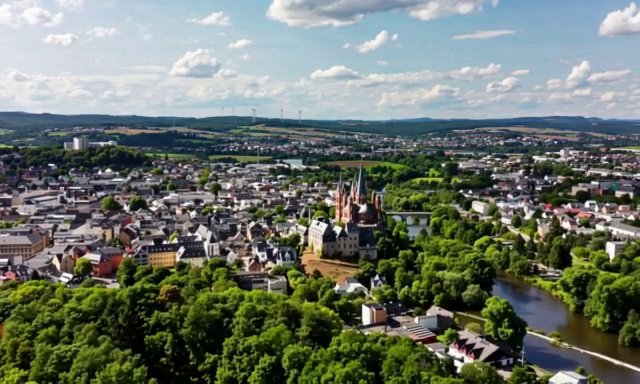} \\[1pt]
      \rowlab{Edited} &
      \includegraphics[width=0.2300\linewidth]{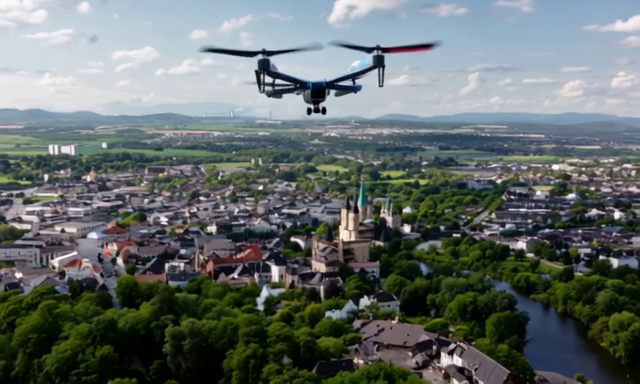}\hfill
      \includegraphics[width=0.2300\linewidth]{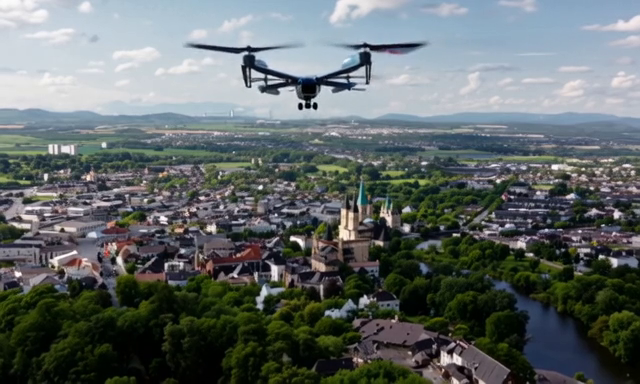}\hfill
      \includegraphics[width=0.2300\linewidth]{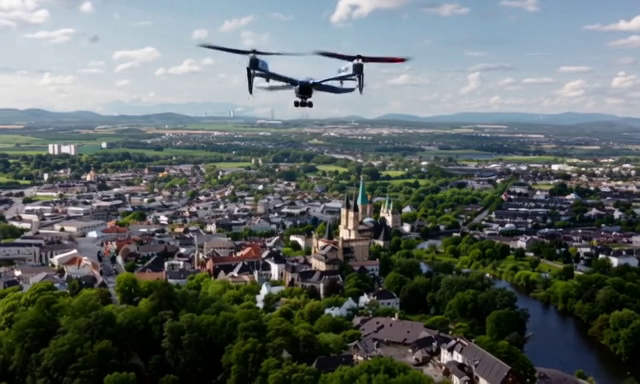}\hfill
      \includegraphics[width=0.2300\linewidth]{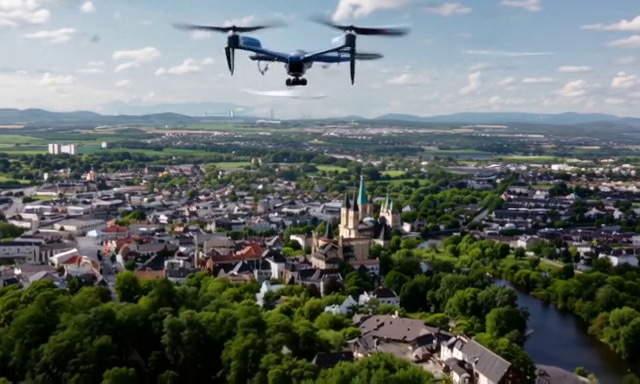}
    \end{tabular}
    \caption{Local addition: \emph{``Add a drone.''}}
  \end{subfigure}\\[2pt]
  \begin{subfigure}{\textwidth}
    \centering
    \begin{tabular}{@{}R@{\hspace{1.5mm}}S@{}}
      \rowlab{Input} &
      \includegraphics[width=0.2300\linewidth]{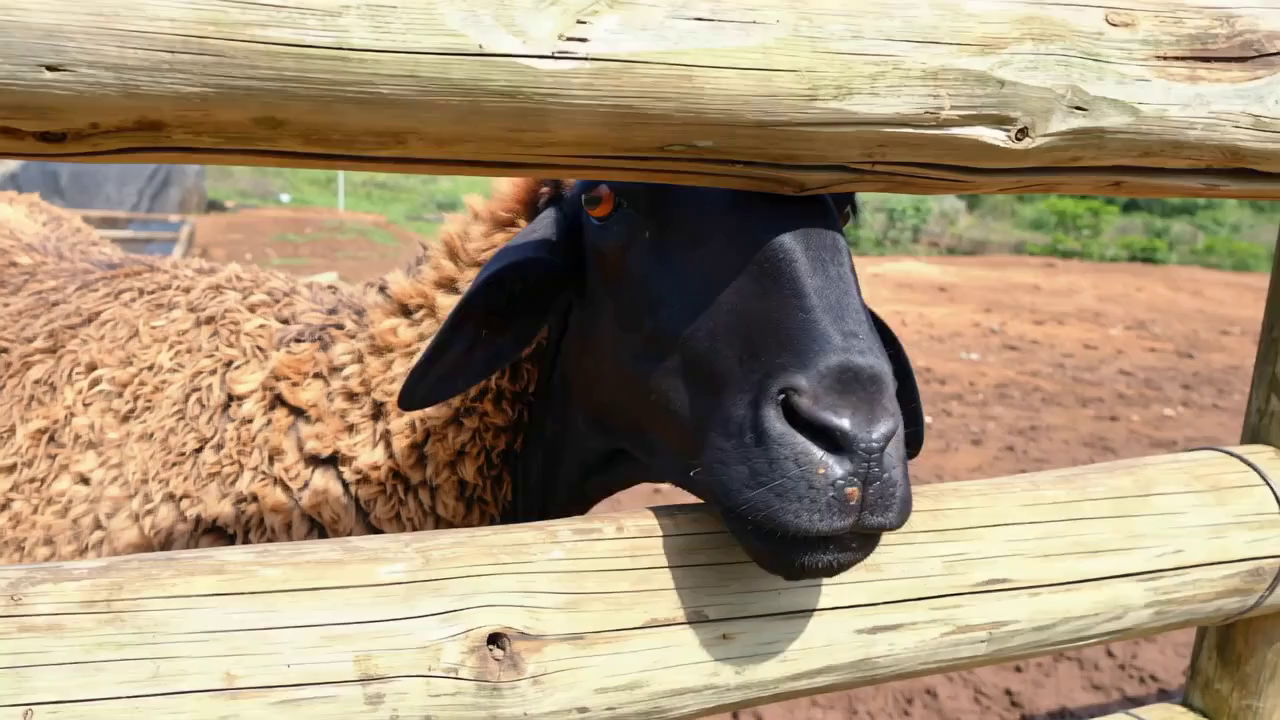}\hfill
      \includegraphics[width=0.2300\linewidth]{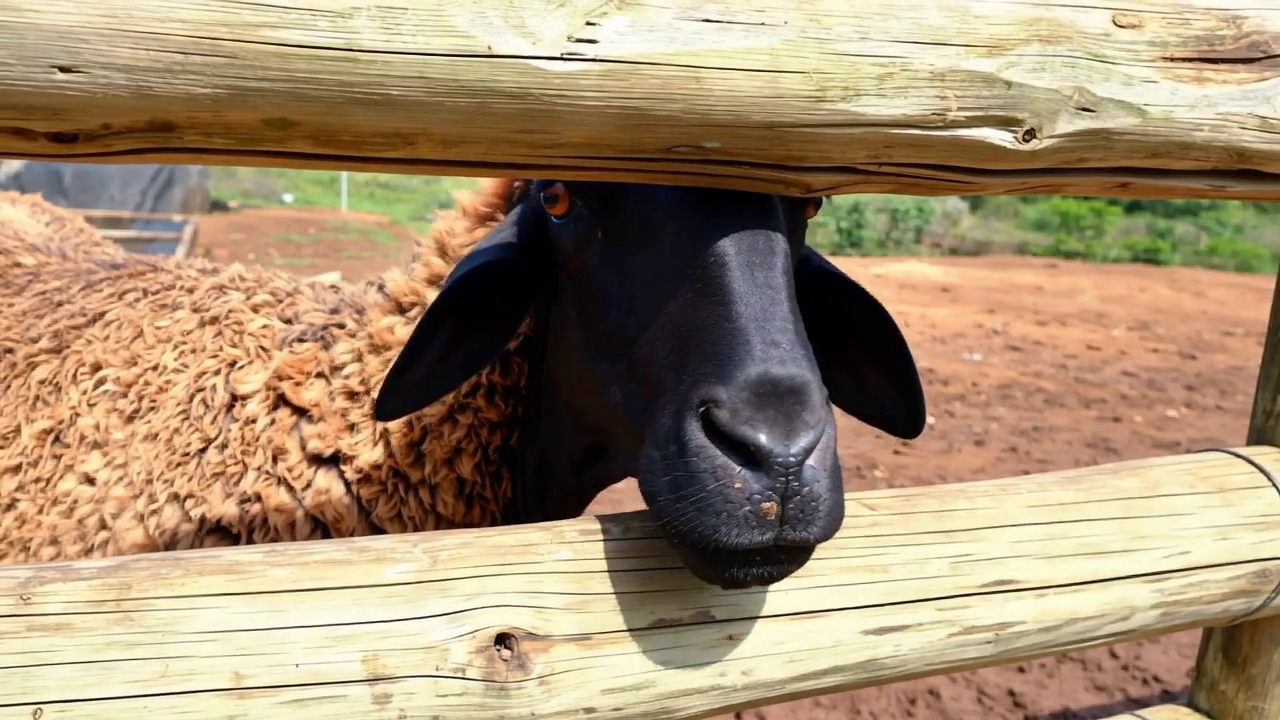}\hfill
      \includegraphics[width=0.2300\linewidth]{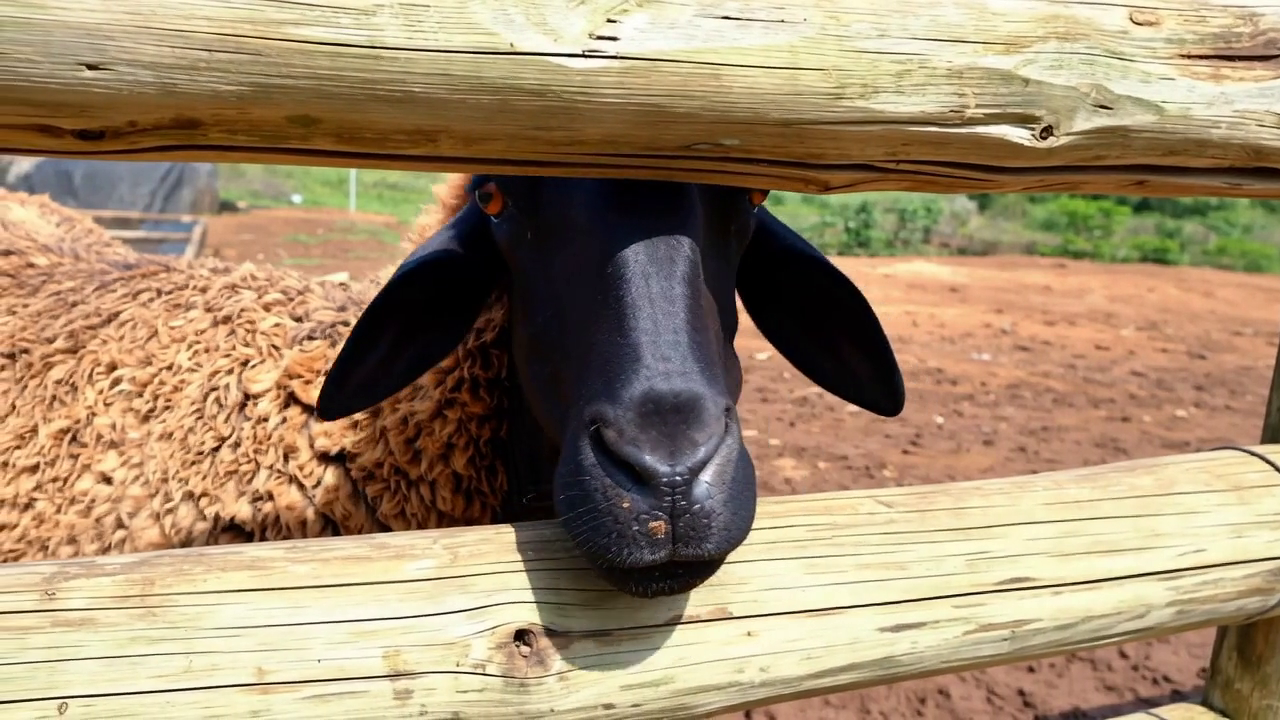}\hfill
      \includegraphics[width=0.2300\linewidth]{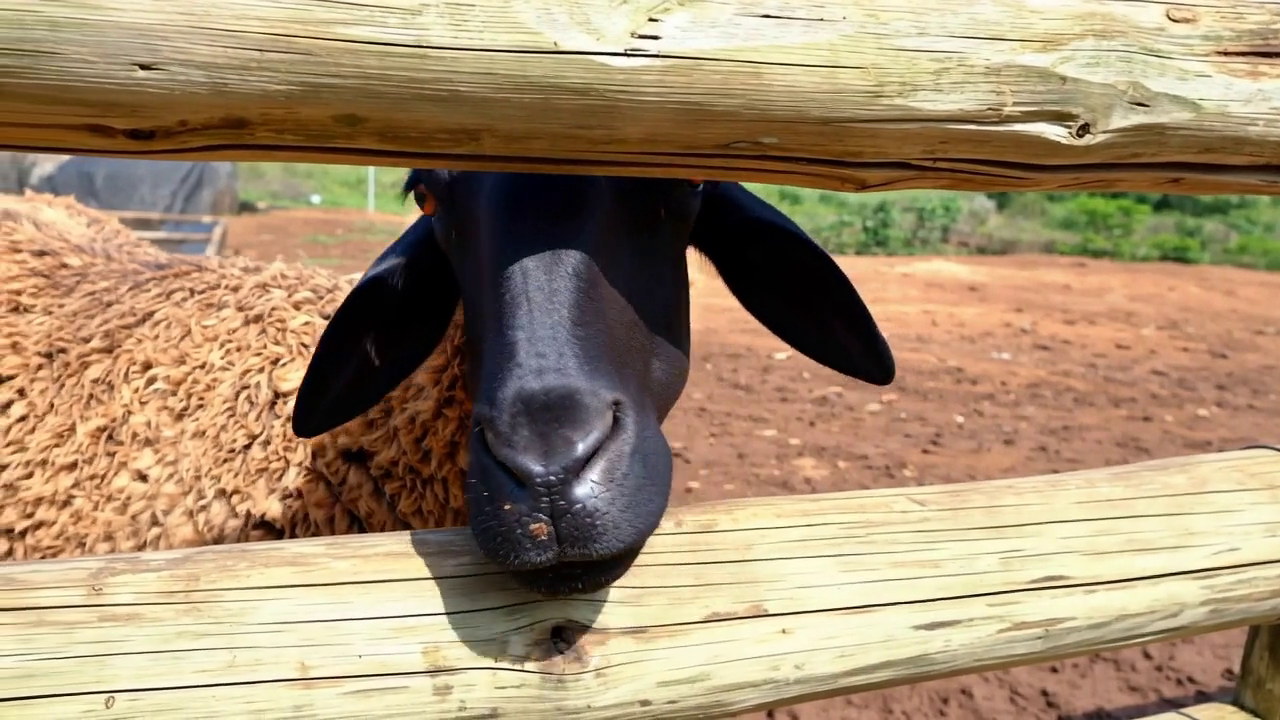} \\[1pt]
      \rowlab{Edited} &
      \includegraphics[width=0.2300\linewidth]{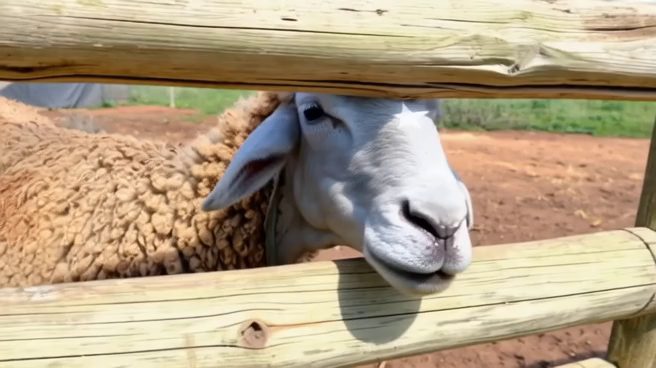}\hfill
      \includegraphics[width=0.2300\linewidth]{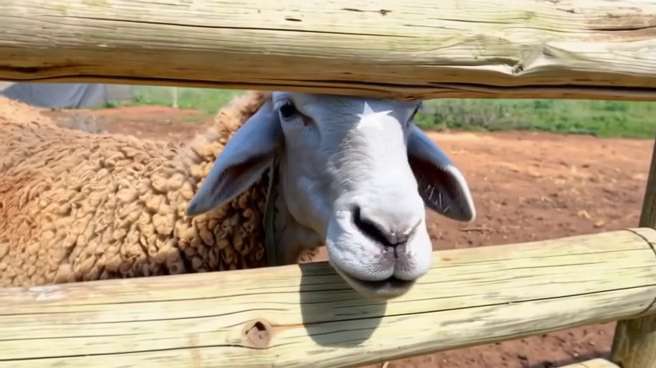}\hfill
      \includegraphics[width=0.2300\linewidth]{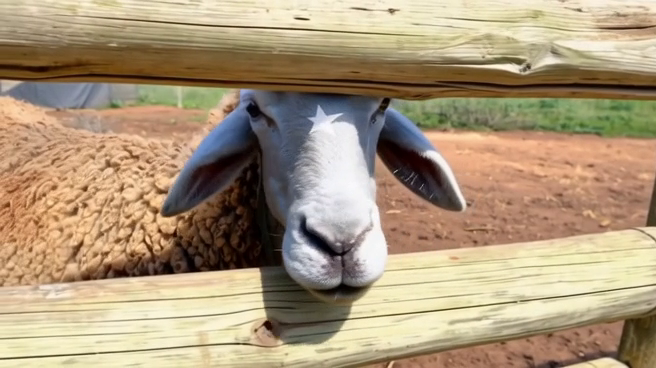}\hfill
      \includegraphics[width=0.2300\linewidth]{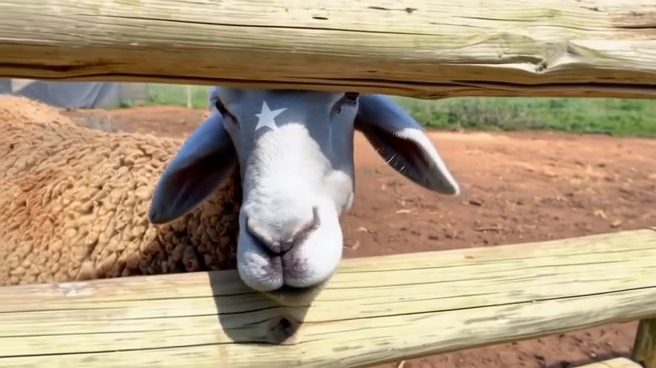}
    \end{tabular}
    \caption{Local swap: \emph{``Replace the dark face of the sheep with a light gray one and add a white star-shaped patch on its forehead.''}}
  \end{subfigure}\\[2pt]
  \begin{subfigure}{\textwidth}
    \centering
    \begin{tabular}{@{}R@{\hspace{1.5mm}}S@{}}
      \rowlab{Input} &
      \includegraphics[width=0.2300\linewidth]{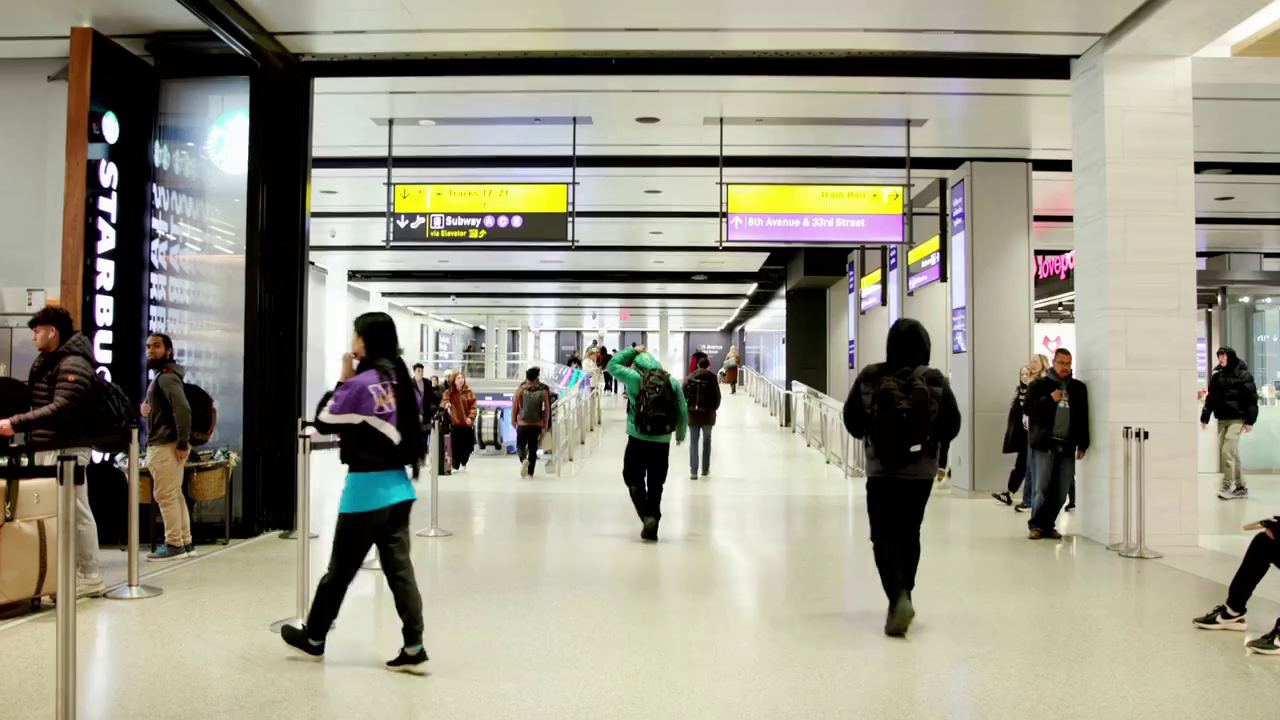}\hfill
      \includegraphics[width=0.2300\linewidth]{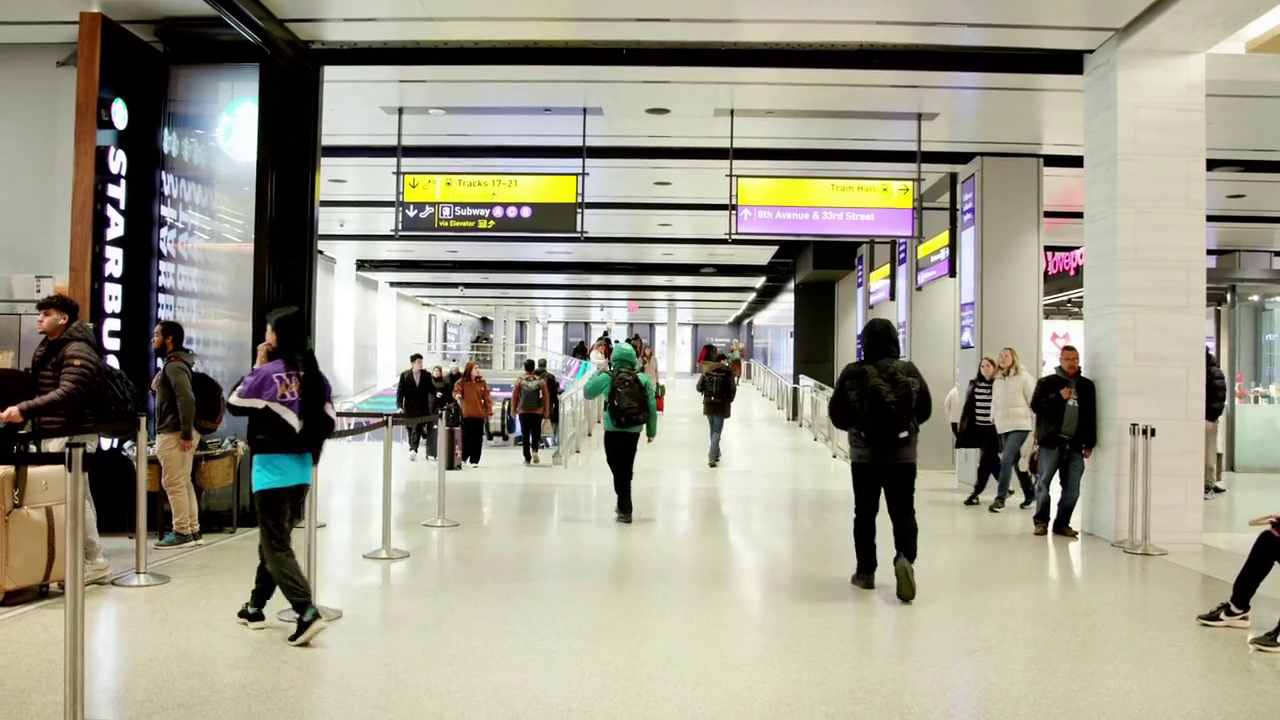}\hfill
      \includegraphics[width=0.2300\linewidth]{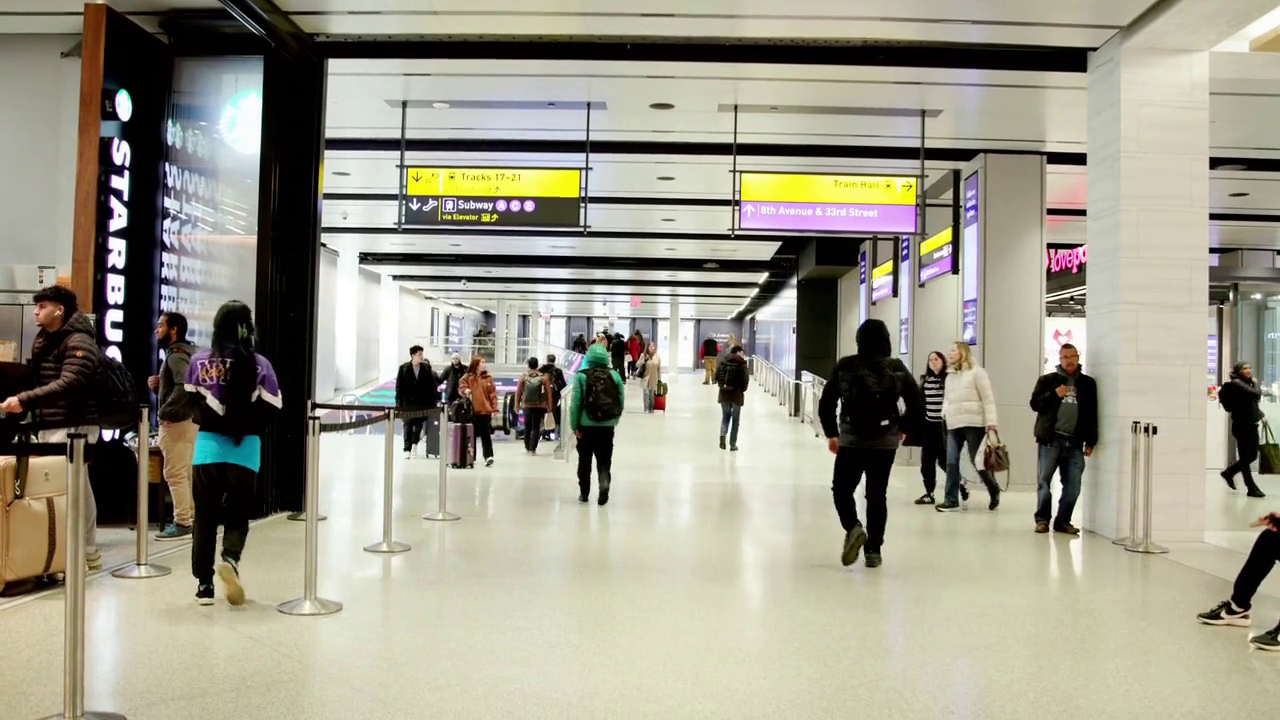}\hfill
      \includegraphics[width=0.2300\linewidth]{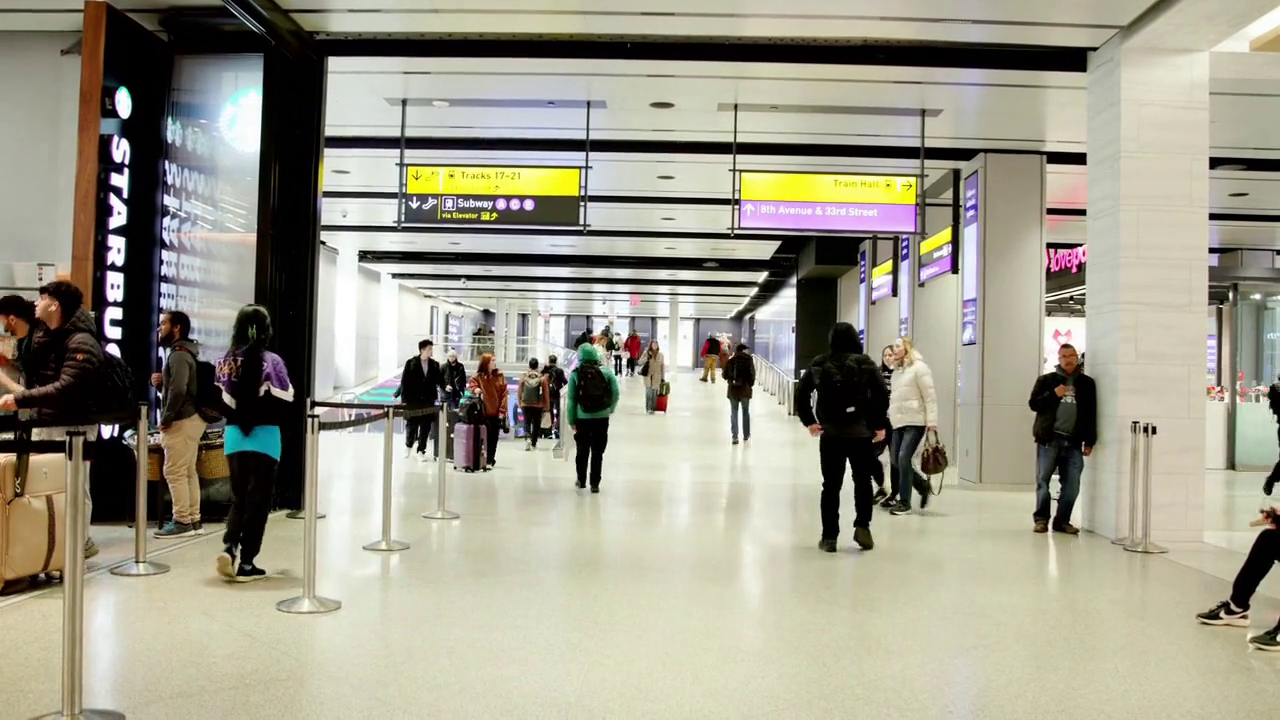} \\[1pt]
      \rowlab{Edited} &
      \includegraphics[width=0.2300\linewidth]{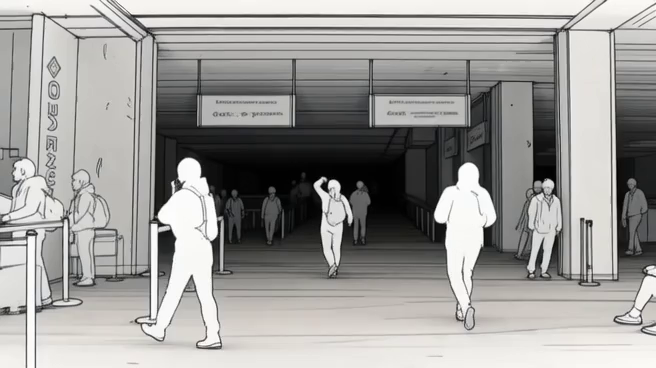}\hfill
      \includegraphics[width=0.2300\linewidth]{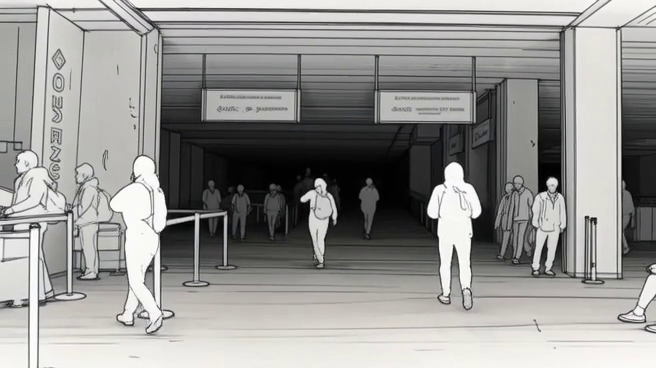}\hfill
      \includegraphics[width=0.2300\linewidth]{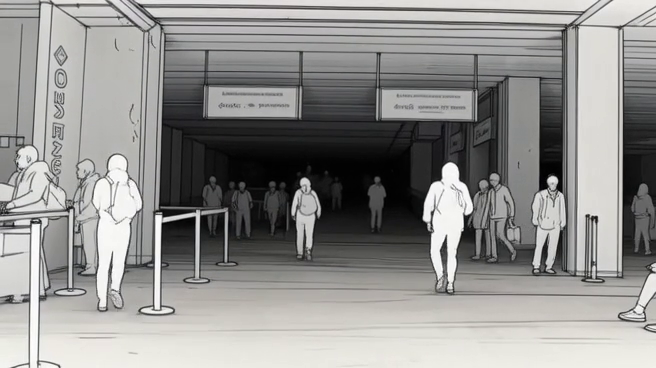}\hfill
      \includegraphics[width=0.2300\linewidth]{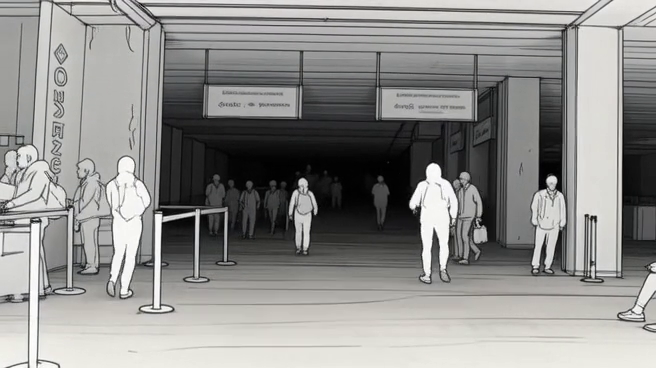}
    \end{tabular}
    \caption{Global stylization: \emph{``Transform the scene into a minimalist monochrome sketch with only lines and shadows, where people are silhouettes, and the signs are abstract symbols that float in space.''}}
  \end{subfigure}
  \caption{\textbf{Qualitative results} across the three main instruction families.}
  \label{fig:results_main}
\end{figure}

\Cref{fig:results_main} shows representative edits across the three main
instruction families: local addition, local attribute swap, and global
stylization. Edits land where the instruction points---the untouched
regions, the camera motion, and the subject motion of the source video are
preserved---and remain stable across frames. \Cref{fig:teaser} additionally
shows a long video edited chunk by chunk with a different instruction per
chunk.

\subsection{Effect of Classifier-Free Guidance}
\label{sec:exp_cfg}

\begin{figure}[!t]
  \centering
  \begin{tabular}{@{}R@{\hspace{1.5mm}}S@{}}
    \rowlab{Input} &
    \includegraphics[width=0.2440\linewidth]{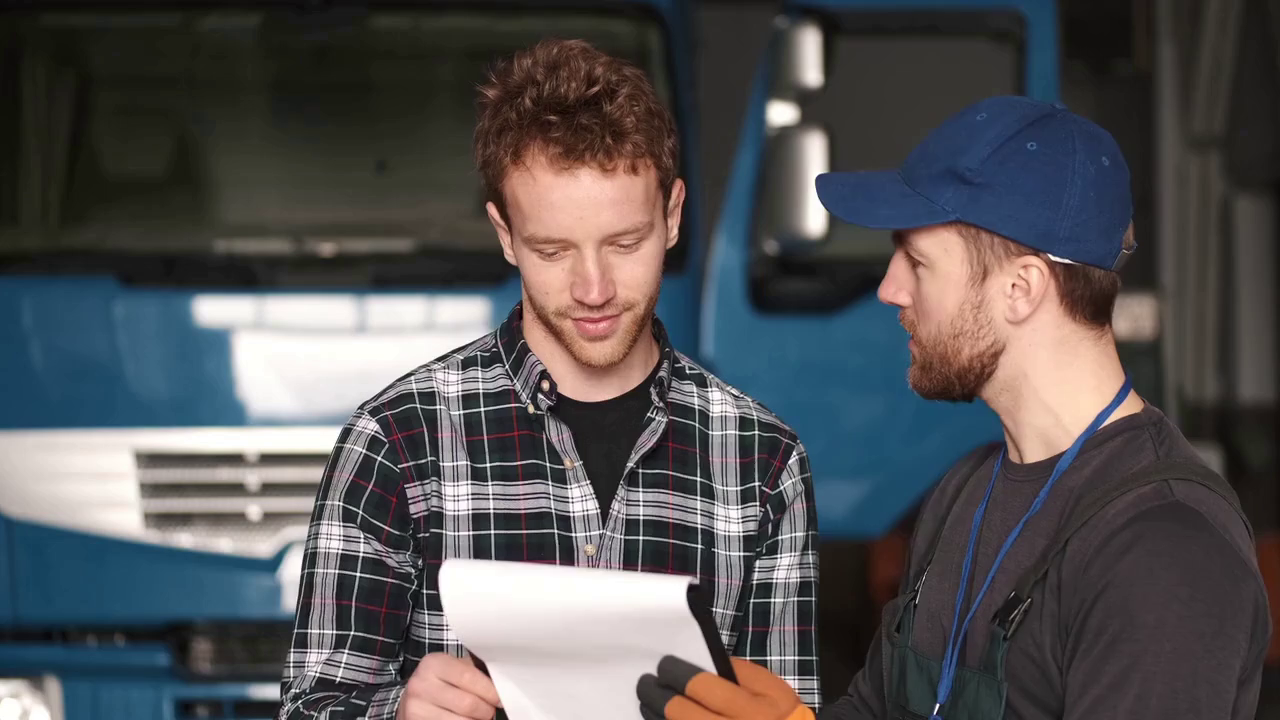}\hfill
    \includegraphics[width=0.2440\linewidth]{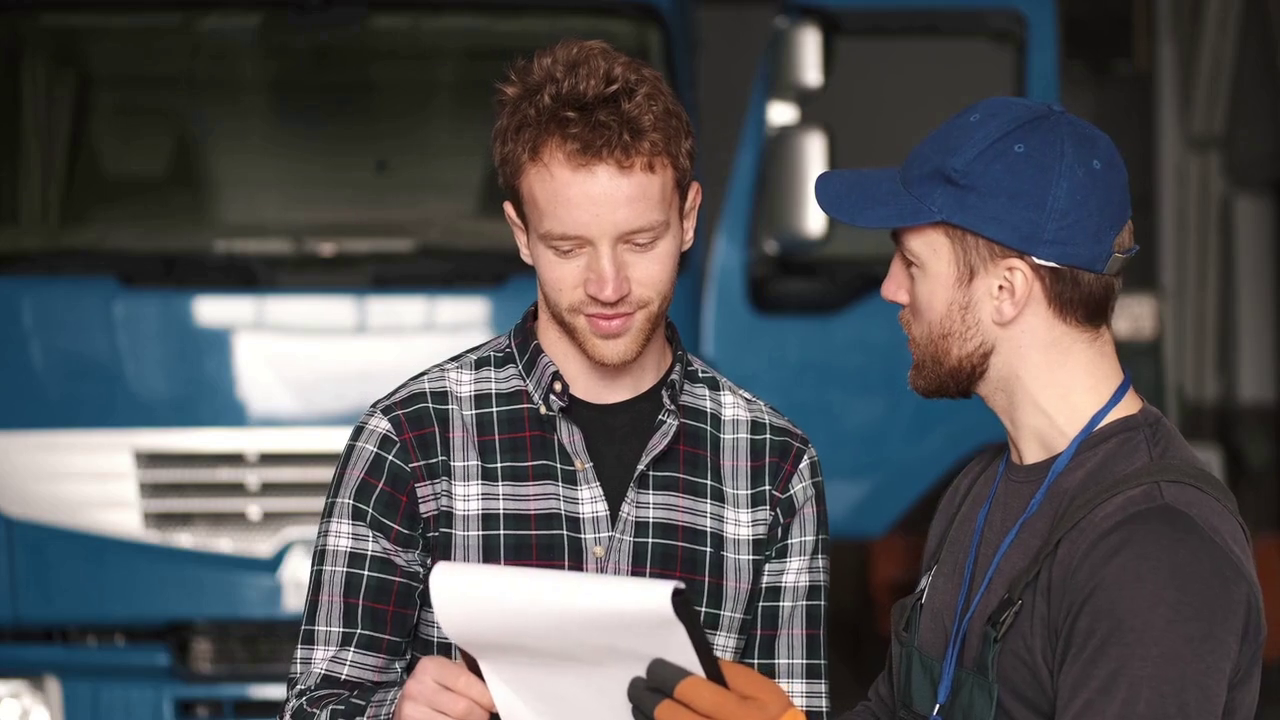}\hfill
    \includegraphics[width=0.2440\linewidth]{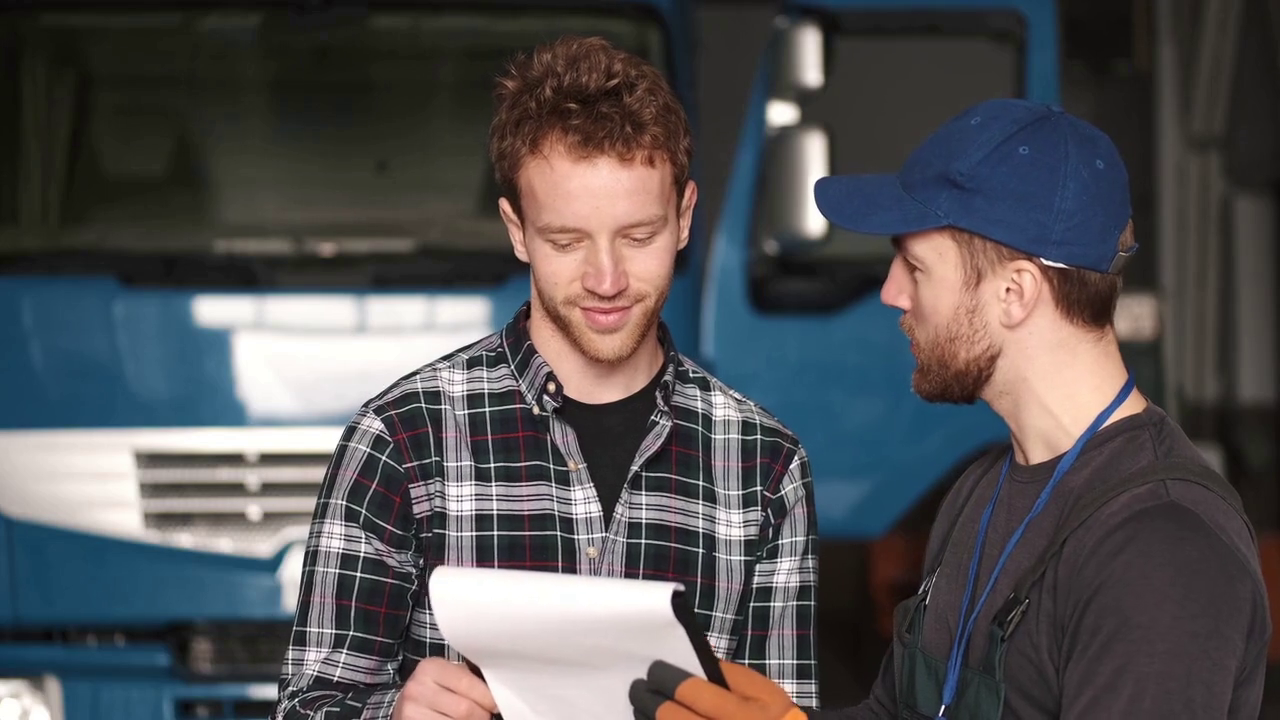}\hfill
    \includegraphics[width=0.2440\linewidth]{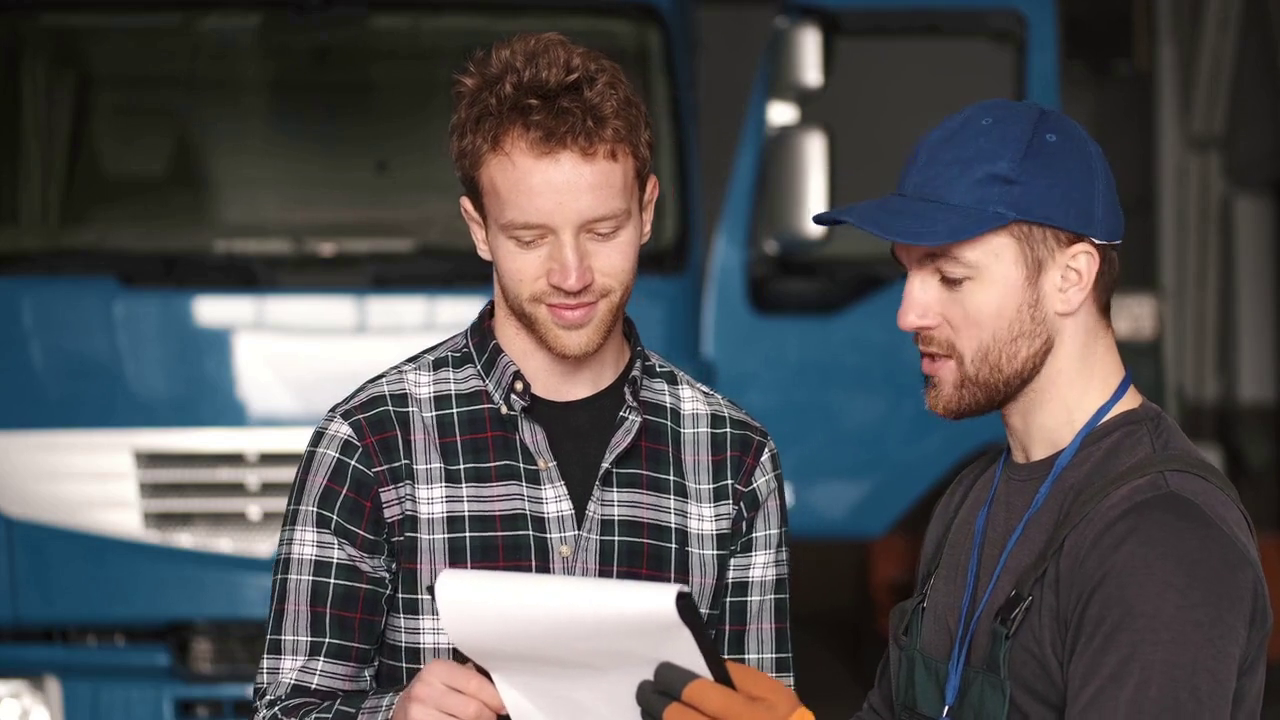} \\[1pt]
    \rowlab{w/o CFG} &
    \includegraphics[width=0.2440\linewidth]{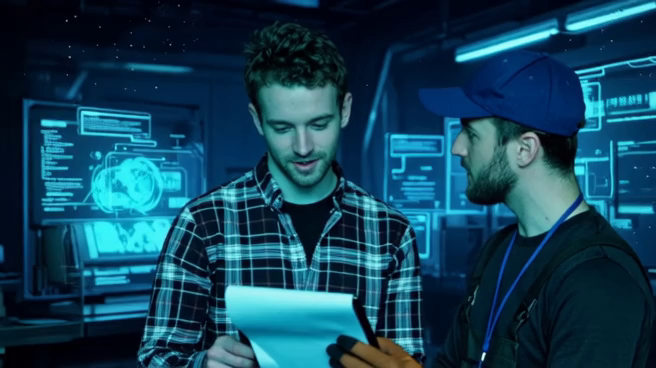}\hfill
    \includegraphics[width=0.2440\linewidth]{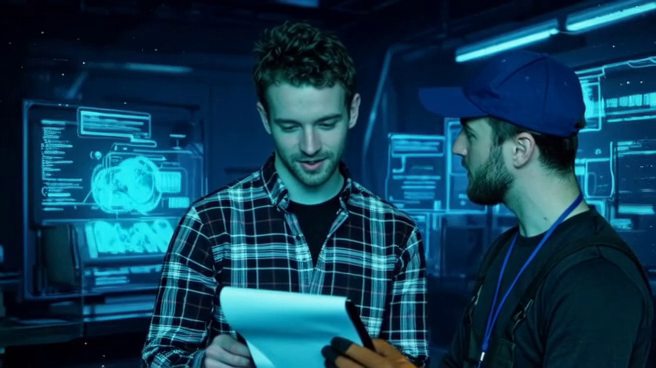}\hfill
    \includegraphics[width=0.2440\linewidth]{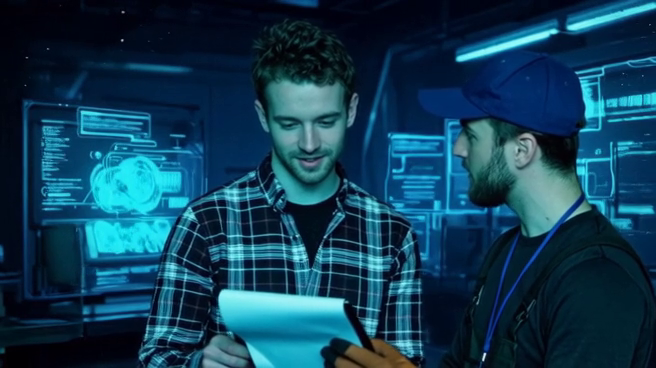}\hfill
    \includegraphics[width=0.2440\linewidth]{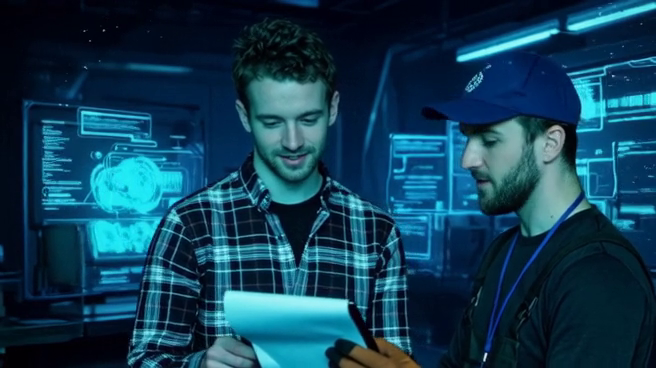} \\[1pt]
    \rowlab{w/ CFG} &
    \includegraphics[width=0.2440\linewidth]{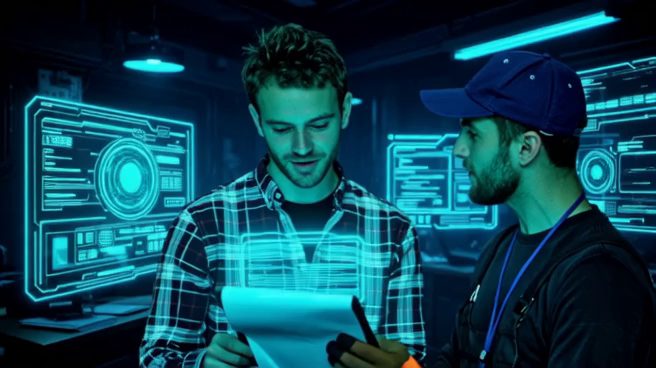}\hfill
    \includegraphics[width=0.2440\linewidth]{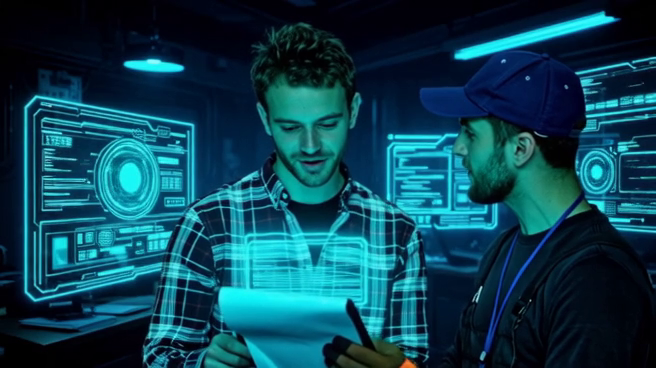}\hfill
    \includegraphics[width=0.2440\linewidth]{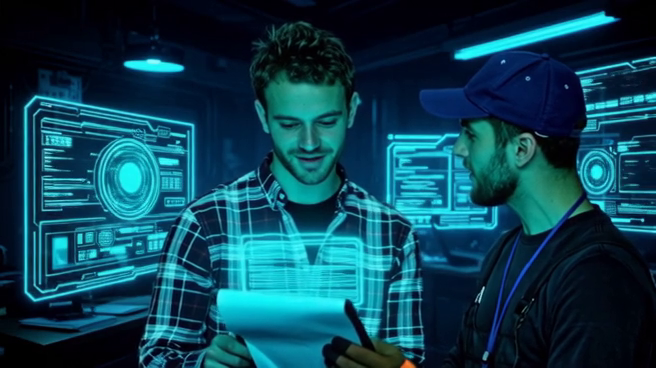}\hfill
    \includegraphics[width=0.2440\linewidth]{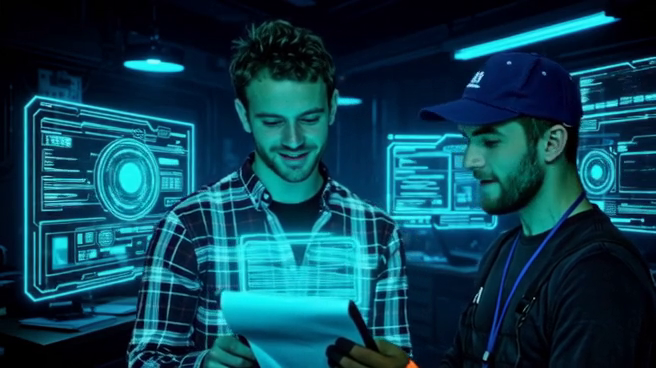}
  \end{tabular}
  \caption{\textbf{Classifier-free guidance ablation} for the instruction \emph{``Transform the scene into a cyberpunk workshop where the documents are holographic, and the men are decoding encrypted data with glowing interfaces.''} Without CFG (middle) the edit is applied only partially; with CFG at scale $4$ (bottom) it is markedly more pronounced.}
  \label{fig:results_cfg}
\end{figure}

Our editing stage uses true classifier-free guidance: a second forward
pass conditioned on an empty negative prompt (and the same reference
latents), combined with the positive prediction at scale $4$ under a
norm-preserving rescale. Since the reference latents alone constrain the
output so strongly, one might expect guidance to be redundant here;
\cref{fig:results_cfg} shows it is not. Without CFG the model tends toward
conservative edits---the instruction is applied only partially, and edited
content stays close to the source. With CFG the edit is markedly more
pronounced and complete, at the cost of a second forward pass per step.
We use CFG scale $4$ by default.

\subsection{Discussion and Limitations}
\label{sec:exp_limitations}

The recipe inherits the image editor's strengths---instruction following,
identity preservation, edit locality---and its blind spots. The main
failure mode we observe is on high-motion content: temporally compressed
Wan latent frames entangle appearance with motion, and the image prior
fits the low-motion regime of that distribution best, so fast or complex
motion can produce softer, less faithful edits than static scenes.

%% file: sections/conclusion.tex
\section{Conclusion}
\label{sec:conclusion}

We presented Qwen-Video-Edit, an instruction-based video editing system
built by repurposing an image editing model. Two warm-started projections,
a grid positional treatment of video latent frames, and a contact-sheet
prompt are sufficient to let Qwen-Image-Edit operate directly on Wan~2.1
video latents; fine-tuning on public data and a few optional Wan~2.2
denoising steps complete the pipeline. No video-pretrained editing
backbone is involved.

Beyond the system itself, we believe the observations behind it are the
more lasting contribution. A great deal of effort goes into training video
VAEs and the latent diffusion models on top of them, and it is tempting to
regard the resulting latent spaces as something qualitatively different
from images. Our experiments suggest otherwise: per-frame video latents
behave, for editing purposes, like modestly re-parameterized image
latents---close enough to the pixel domain that a mature image prior
transfers across with a linear-scale bridge, and the residual gap is
concentrated exactly where temporal compression entangles appearance with
motion. Put plainly, the latent space is not that different from pixel
space. We hope this encourages more work that treats strong image models
as first-class citizens for video tasks, and that reserves video-specific
training capacity for what is genuinely temporal.